\documentclass[preprints,article,accept,moreauthors,pdftex]{Definitions/mdpi} 

\firstpage{1} 
\pubvolume{1}
\issuenum{1}
\articlenumber{0}
\pubyear{2021}
\copyrightyear{2021}
\datereceived{} 
\dateaccepted{} 
\datepublished{} 
\hreflink{https://doi.org/} 
\pdfoutput=1

\usepackage{adjustbox} 
\definecolor{Gray}{gray}{0.85}
\usepackage{subcaption} 
\usepackage{dsfont} 

\usepackage{todonotes}

\newcommand{\h}{\cellcolor{black!20}\bfseries}

\Title{A Statistics and Deep Learning Hybrid Method for Multivariate Time Series Forecasting and Mortality Modeling}

\TitleCitation{Title}

\Author{Thabang Mathonsi $^{1,\dagger, \ddagger}$\orcidA{} and Terence L. van Zyl $^{2,\dagger}$\orcidB{}}

\AuthorNames{Thabang Mathonsi and Terence L. van Zyl}

\AuthorCitation{Mathonsi, T.; van Zyl, T. L.}

\address{%
	$^{1}$ \quad School of Computer Science and Applied Mathematics, \emph{University of the Witwatersrand}, Johannesburg, South Africa; thabang.mathonsi@wits.ac.za\\
	$^{2}$ \quad Institute for Intelligent Systems, \emph{University of Johannesburg}, South Africa; tvanzyl@uj.ac.za}

\corres{Correspondence: thabang.mathonsi@wits.ac.za}

\firstnote{These authors contributed equally to the conceptualisation and presentation of this work.} 
\secondnote{These authors implemented the methods and executed the empirical experiments.}

\abstract{Hybrid methods have been shown to outperform pure statistical and pure deep learning methods at forecasting tasks and quantifying the associated uncertainty with those forecasts (prediction intervals). One example is Exponential Smoothing Recurrent Neural Network (ES-RNN), a hybrid between a statistical forecasting model and a recurrent neural network variant. ES-RNN achieves a 9.4\% improvement in absolute error in the Makridakis-4 Forecasting Competition. This improvement and similar outperformance from other hybrid models have primarily been demonstrated only on univariate datasets. Difficulties with applying hybrid forecast methods to multivariate data include ($i$) the high computational cost involved in hyperparameter tuning for models that are not parsimonious, ($ii$) challenges associated with auto-correlation inherent in the data, as well as ($iii$) complex dependency (cross-correlation) between the covariates that may be hard to capture. This paper presents Multivariate Exponential Smoothing Long Short Term Memory (MES-LSTM), a generalized multivariate extension to ES-RNN, that overcomes these challenges. MES-LSTM utilizes a vectorized implementation. We test MES-LSTM on several aggregated coronavirus disease of 2019 (COVID-19) morbidity datasets and find our hybrid approach shows consistent, significant improvement over pure statistical and deep learning methods at forecast accuracy and prediction interval construction.}

\keyword{deep learning; multivariate time series forecasting; prediction intervals; mortality modeling  } 

\begin{document}
	
	\iftrue
\end{paracol}
\fi
\begin{paracol}{2}
	\switchcolumn
	
	
	\section{Introduction}
	Morbidity and mortality modeling is crucial for planning in global economies, national healthcare systems, and other industries such as insurance. Practitioners from statistics, machine learning, and actuarial backgrounds have invested into improving the accuracy of morbidity and mortality forecasting. Some recent advances have emerged from the fields of hybrid models, and interpretable models such as Temporal Fusion Transformers~\cite{Lim21}. Despite these advances, the recent devastating global impact of the novel coronavirus disease of 2019 (COVID-19) virus has highlighted the importance of effective planning from government agencies and healthcare bodies across the globe. This kind of planning requires reliable projections into the future, and as a result there exists a need for improved forecasting techniques and methods.
	
	Smyl~\cite{Smyl20} developed a hybrid method for generating point forecasts and quantifying the uncertainty associated with those point forecasts. The model quantifies uncertainty by producing prediction intervals at the 1\%, 5\% and 10\% levels of significance. The hybrid method combines Exponential Smoothing (ES, \cite{Hynd02}) with Recurrent Neural Networks (RNN, \cite{Jaeger01}) and the resulting scheme is referred to as ES-RNN. This name is a bit of a misnomer as Smyl's~\cite{Smyl20} methodology actually involves combining ES, with a varaiant of RNN called Long Short-Term Memory (LSTM, \cite{Hoch97, Hoch01}).
	
	Smyl~\cite{Smyl20} purports that their method produces "...forecasts that are more accurate than those generated by either pure statistical or pure [machine learning] approaches, thus exploiting their advantages while avoiding their drawbacks." The hybrid method, ES-RNN, outperformed pure statistical and pure deep learning methods submitted alongside it to the Makridakis-4 (M4) Forecasting Competition~\cite{Makri20a}. The M4 Competition provides to Competition participants 100000 datasets from a cross-section of industries and applications for the entrants to apply their techniques to.
	
	\citet{Red19} extend the Smyl ES-RNN~\cite{Smyl20} hybrid technique by executing a graphical processing unit (GPU) implementation (GPU-ES-RNN). Using a vectorized and GU-based implementation of the original ES-RNN,~\citet{Red19} state they achieve up to 322x increase in training speed, largely attributed to batching and parallelization. The authors report that their network produces performance results similar to those reported in the original Smyl submission. Furthermore, the GPU implementation migrates the original C++ code to Python and PyTorch.
	
	In both cases (ES-RNN and GPU-ES-RNN), the authors focus on univariate datasets. However, a significant number of real world forecasting involves multivariate datasets~\cite{Beer21}. In some cases, researchers have found it useful to augment their univariate models with multivariate exogenous factors and/or multivariate analogues of their models to improve predictive accuracy~\cite{Bhar21}. In addition, the complexities associated with modeling morbidity and mortality often require multiple data sources in order to fit a model that better describes the real world situation~\cite{Olk01}.
	
	This paper extends both the ES-RNN and GPU-ES-RNN research by adapting the ES and LSTM hybrid method to make it applicable to the multivariate case. By incorporating exogenous factors throughout, our extension departs from the classical univariate case and, as such, poses more complications associated with, for instance, auto-correlation inherent in the data and cross-correlation dependencies between covariates.
	
	Our model incorporates exogenous covariates through the use of global and local variables, i.e. those derived from all the data or large segments of it, and those derived from separate covariates. This combination allows the model to cross-learn at the same time leverage information presented at a granular level.
	
	Essentially, we have three workstreams i.e. multivariate exponential smoothing, recurrent neural networks in general and LSTM in particular, and hybrid methods. We review all three workstreams next.
	
	\subsection{Literature Review}
	\label{sub:litreview}
	There are several notable works concerning multivariate extensions of exponential smoothing.~\citet{Jon66} applies recursion for estimating the smoothing matrix.~\citet{Enn82} consider a class of various exponential smoothing models and use them as a proxy for the univariate exponential smoothing couterpart.~\citet{Tri67} adapt the smoothing matrix of their so-called adaptive models periodically, with the aid of maximum likelihood estimation.~\citet{Hav86} further simplifies the class of models proposed by~\citet{Enn82} and the simplifications enable the use of univariate smoothing models in forecasting tasks for correlated time series. Moreover, Harvey's~\cite{Hav86} results are also valid in the case where the smoothing models exhibit polynomial trend and seasonal components.~\citet{Pfe83} focus on structural models aimed at producing optimal forecasts. By building upon previous multivariate time series exponential smoothing research,~\citet{Pfe83} also offer detailed instructions for parameter initialization and model-fitting.
	
	Harvey's approach is most suitable for our interests for two reasons. One, because of reduced computational complexity; and two, we find that adapting Harvey's approach yields similar results to using an adapted version of, for example, the more complex~\citet{Pfe83} approach. The second reason is due to the LSTM's ability to model complex interdependency between covariates~\cite{Tan21, Hu21}, which negates the need to model the cross-correlation using the preprocessing layer statistical methods.
	
	The recent COVID-19 pandemic has provided a unique opportunity due to ($i$) a multitude of open datasets and ($ii$) extensive related research exploring the performance of multivariate forecasting models. In the domain of COVID-19-related research, LSTM networks have been applied by, for instance,~\citet{Ism20}, who focus their attention on the spread of the pandemic in countries including Denmark, Belgium, Germany, France, United Kingdom, Finland, Switzerland and Turkey.~\citet{Ism20} model using Auto-Regressive Integrated Moving Average (ARIMA) and Nonlinear Autoregression Neural Network (NARNN, \cite{Ibra16}) as benchmarks and report that LSTM has superior accuracy in terms of forecasting the cumulative cases of infected individuals.
	
	\citet{Roh21} use variations of LSTM including Bidirectional LSTM (Bi-LSTM), and encoder-decoder LSTM (ed-LSTM) models for multi-step intra-country forecasting in the short-term in India. They cite challenges in the modeling process caused by difficulties in capturing, in any available COVID-19 dataset, potentially important multivariate factors such as population density, travel logistics, and other societal issues (e.g. the general standard of living). If they were available, these exogenous factors could be useful in improving predictive skill in the modeling process.
	
	\citet{Chim20} base their forecasts on training data that they acquired from the John Hopkins University and the Canadian Health Authority. They use a standard implementation of LSTM. They forecast the pandemic ending in June 2020 (which we now know to be incorrect).
	
	\citet{Shah20} model the rises and declines of confirmed cases, deaths and recoveries in ten countries, including China. They rate their model performances from best to poorest as Bi-LSTM, LSTM, Gated Recurrent Units (GRU,~\cite{Chung14}), Support Vector Regressor (SVR) and ARIMA. They report Bi-LSTM has superior performance over the other models with the lowest MAE and RMSE values of 0.0070 and 0.0077, respectively.
	
	As evidenced by previous related research, there are successes and failures with using LSTM to forecast COVID-19. Various factors contribute to the at times the poor performance of LSTM and its variations in this regard. One such factor is that the pandemic is relatively recent, and any available dataset is a small fraction of the requisite volume of training data the data-hungry artificial neural networks require. LSTM has been shown to perform well in a variety of applications with datasets that are sufficiently long~\cite{Mat2}. In our research, we circumvent the problem of limited data by integrating a small, parsimonious LSTM network in our model. We offer more details about the structure of our model in Section~\ref{sub:hyperparams}.
	
	Univariate forecasting with the aid of pure machine learning has been conducted since as far back as the works of~\citet{Hu64}. Techniques merging machine learning with statistical methods have since gained popularity.
	
	Recently, the original ES-RNN hybrid technique won over hundreds of other submissions both in terms of ($i$) point forecasts as well as ($ii$) their associated prediction intervals. Both components are difficult to model, but perhaps the latter even more so. Deep learning models often do not have a mechanism for quantifying uncertainty~\cite{Mat1}. In such cases, prediction intervals must be constructed using computational mechanisms~\cite{Mat2}.
	
	\citet{Bor20}, however, produce forecast machinery comprised of deep, fully connected layers and report that their model outperforms ES-RNN on the M4 Competition dataset. Based on Neural Basis Expansion Analysis (NBEATS),~\citet{Bor20} conclude that pure deep learning is better than hybrid methods as their method outperforms the best statistical and the best hybrid method by 11\% and 3\%, respectively. The technique has since been extended to include exogenous factors (NBEATS-x,~\cite{Kin21}).
	
	Further evidence has emerged since the end of the Makridakis-5 (M5) Accuracy and Uncertainty Competitions~\cite{Makri21a, Makri21b} that pure deep learning models may be superior for hierarchical forecasting. However, hybrid methods are still worth investigating in the multivariate with exogenous variables forecasting setting. Starting here as a point of departure, we hypothesise that a hybrid technique such as ES-RNN may work just as well in this setting (exploiting a simple LSTM’s ability to model cross-correlation).
	
	Furthermore, extending the current ES-RNN hybrid forecasting research to the multivariate case is crucial as multivariate datasets are usually more representative of realistic forecasting scenarios likely to be encountered in other applications.
	
	\subsection{Contribution}
	Our contribution can be summarized as follows:
	\begin{itemize}
		\item We extend the current research, which focuses on ES and RNN hybrid methods for univariate forecasting, to a multivariate framework. We thus test assertions on multivariate mortality data with exogenous variables previously only tested empirically on univariate data, i.e. are hybrid methods better than pure statistical or pure deep learning methods at ($i$) forecasting tasks and ($ii$) quantifying forecast uncertainty? In particular, we present a natural extension of Smyl's ES-RNN to higher dimensions; 
		\item we present our forecast engine MES-LSTM, which is an efficient generalization, and as such, may be applied not only to the multivariate case but also to the univariate setting with ease; and
		\item whereas previous (univariate) research on forecasting hybrid methods primarily focuses on the multiplicative seasonality case, we consider both multiplicative and additive, with automatic adaptation to the case most applicable to the particular dataset.
	\end{itemize}
	
	The remainder of this paper is organized as follows. In Section~\ref{sec:method} we describe our architecture in detail, and how we merge the classical state-space forecast model with the advanced artificial neural network. We also describe how we evaluate our model's performance. In Section~\ref{sec:data} we describe the datasets used. The data section is followed by Section~\ref{sec:results}, where we  discuss our results in detail. We conclude in Section~\ref{sec:conclusion} with a few key points and possibilities for extending this research to future work.
	
	\section{Methods}
	\label{sec:method}
	We employ GPU computation by utilising Tensorflow's eager execution to transform the original Smyl~\cite{Smyl20} ES-RNN to a generalised multivariate implementation. In their GPU implementation,~\citet{Red19} initialise per-series attributes according to the guidelines of the M4 Competition. We initialise per-covariate parameters, and have a vectorized ES-LSTM, which we term Multivariate ES-LSTM (MES-LSTM).
	
	The other difference between ES-RNN and MES-LSTM is that we use the most suitable between additive or multiplicative seasonality, whereas previous authors have only considered multiplicative seasonality. The most suitable seasonality structure in each case is ascertained by tracking the exponential smoothing fit on the training data.
	
	In the remainder of this chapter, we explicitly define the quintessential aspects of MES-LSTM.  The description is analogous to that of Smyl~\cite{Smyl20}, with relevant extensions made to suit our multivariate presentation. Our model comprises two distinct layers: an exponential smoothing layer inside our preprocessing module, and an LSTM layer used for learning the dependency between the covariates. This methodology is consistent with Section~\ref{sub:litreview}, where we outline our choice to expand upon Harvey's approach~\cite{Hav86} over the~\citet{Pfe83} approach for multivariate exponential smoothing.
	
	Concisely, our model learns the parameters associated with each covariate in the preprocessing step, then learns the correlation between the covariates in the deep learning layer. Parameters optimized in both steps are then used at the inference stage, where the prediction of multivariate point forecasts are produced, and the uncertainty associated with these forecasts is quantified.
	
	\subsection{Preprocessing layer}
	\label{subsec:prelayer}
	For each covariate in our multivariate sample space, let $\left\{ {y} \right\}_t$ represent, for example, a weekly time series (weekly series for ease of exposition only) which we assume can be decomposed into the additive form
	\begin{align}
		y_t &= l_t + b_t + s_t + \varepsilon_t, 
	\end{align}
	where $l_t$ is the level in week $t$, $s_t$ the seasonal effect and $\varepsilon_t$ the noise term centered at zero with constant variance. Let $\hat{l}_t$ denote the estimated level in week $t$ and $\hat{b}_t$ the corresponding trend estimate. Let the estimate of the seasonal effect at time $t$ corresponding to week $t + i, \,i = 1, \,2, \,\dots, \,52$ be denoted by $\hat{s}_{t, t + i}$. The estimates ${\hat{s}_{t, t + i}}$ satisfy the condition $\sum_{i = 1}^{52} \hat{s}_{t, t + i} = 0$. When a new observation $y_{t+1}$ becomes available, all estimated components, i.e. seasonality, trend, and level, are updated with the aid of three smoothing constants $0 \leq \alpha, \,\gamma, \,\delta \leq 1$ as follows:
	\begin{align}
		\hat{l}_{t +1} &= \alpha \left(y_{t + 1} - \hat{s}_{t, t + 1} \right) + \left(1 - \alpha \right) \left( \hat{l}_t + \hat{b}_{t} \right) \label{eq:level}\\
		\hat{b}_{t+ 1} &= \gamma \left( \hat{l}_{t + 1} - \hat{l}_t\right) + \left(1 - \gamma \right) \hat{b}_t \label{eq:trend} \\ 
		\hat{s}^*_{t + 1, t + 1} &= \delta \left(y_{t + 1} - \hat{l}_{t + 1} \right) + \left( 1 - \delta \right)\hat{s}_{t, t + 1} \label{eq:season}\\
		\sum_{i = 0}^{51} \hat{s}^*_{t + 1, t + 1 + i} &= 0. \label{eq:standard}
	\end{align}
	Equations \eqref{eq:season} and \eqref{eq:standard} define a two-step computational procedure of the seasonal effects ${\hat{s}_{t + 1, \,t + 1 + i}}$. The second step standardizes the initial values computed in the first (with the asterisk indicating "intermediary value") so that they sum to zero.
	
	The forecast at time $t$ of a future out-of-sample realization ($m>i$) $y_{t + m}$ is given by
	\begin{align}
		\hat{y}_{t, t + m} &= \hat{l}_t + m\hat{b}_t + \hat{s}_{t, t + m}\label{eq:pred}.
	\end{align}
	For the multiplicative seasonality case, make the substitution $\left\{y \right\}_t = \ln \left( \left\{ y \right\}_t \right)$. This substitution amounts to assuming the original series level changes in approximately constant rates instead of constant increments. The substitution also requires changing the condition imposed on the estimates of the multiplicative seasonal factors over the 52 weeks from summing to zero to their geometric mean equalling one. Equations \eqref{eq:level} to \eqref{eq:standard} then become
	\begin{align}
		\hat{l}_{t + 1} &= \alpha(\frac{y_{t + 1}}{\hat{s}_{t,t + 1}}) + (1 - \alpha)\hat{l}_{t} \hat{b}_{t} \label{eq:level1}\\
		\hat{b}_{t + 1} &= \gamma(\frac{\hat{l}_{t + 1}}{\hat{l}_{t}}) + (1- \gamma) \hat{b}_{t} \label{eq:trend1} \\ 
		\hat{s}^*_{t + 1, t + 1} &= \delta \frac{y_{t + 1}}{\hat{l}_{t} \hat{b}_{t}} + (1-\delta)\hat{s}_{t, t + 1} \label{eq:season1}\\
		\left(\prod_{i = 0}^{51} \hat{s}^*_{t + 1, t + 1 + i}\right)^{\frac{1}{51}} &= 1,
	\end{align}
	and the predictive Equation \eqref{eq:pred} becomes
	\begin{align}
		\hat{y}_{t, t + m} &= \hat{l}_t \left(\hat{b}_t \right)^m \hat{s}_{t, t + m}. \label{eq:pred1}
	\end{align}
	In both the additive and multiplicative seasonality cases expressed above, the initial smoothing parameters are estimated as follows. The level is taken to be the overall in-sample average. The trend is initialized as the difference between the first and last in-sample observations, divided by the total number of increments. Seasonality is initialized by the deviations between the first week's observations and the level plus trend fit. More details about the dimensionality of the parameter space are given in Section~\ref{sub:hyperparams}.
	
	Now that we have the formulations for both the additive and multiplicative seasonality cases, we can merge the preprocessing layer with our deep learning component as follows. For simplicity, we keep the input size and the size of the output predictions equal.
	
	For the univariate additive case we have
	\begin{align}
		\hat{y}_{t,~t + 1, \,\dots,\,t + m} &= \hat{l}_t + [t + 1,~...,~t + m]^\top \odot \text{LSTM}(X_t) + \hat{s}_{t,~t + 1,~\dots,~t + m},
		\label{eq:modified}
	\end{align}
	where $\odot$ is the Hadamard product, and $X_t$ denotes a vector of de-trended and de-seasonalized observations of which a scalar element $x_i$ is given by:
	\begin{align}
		x_i &= y_i - \hat{l}_i - \hat{s}_i,~\textnormal{for } i = t + 1,~\dots,~t + m. \label{eq:x}
	\end{align}
	
	For the multivariate additive case we have 
	\begin{align}
		\hat{\boldsymbol{y}}_{t,~t + 1,~\dots,~t + m} &= \hat{\boldsymbol{l}}_t + [t + 1,~...,~t + m]^\top \odot \text{LSTM}(\boldsymbol{X}_t) + \hat{\boldsymbol{s}}_{t,~t + 1,~\dots,~t + m}, \label{eq:mmodified}
	\end{align}
	where (if we have, say, $k$ covariates in total) $\boldsymbol{X}_t$ is a $k$ x $m$ matrix of de-trended, de-seasonalized features of which a vector component (for each covariate) $\boldsymbol{x}_i$ is calculated via the equation: 
	\begin{align}
		\boldsymbol{x}_i &= \boldsymbol{y}_i - \hat{\boldsymbol{l}}_i - \hat{\boldsymbol{s}}_i,~\text{for } i = t + 1,~\dots,~t + m.
		\label{eq:mx}
	\end{align}
	Here, the vectors $\hat{\boldsymbol{l}}_i$ and $\hat{\boldsymbol{s}}_i$ are the same dimension as $\boldsymbol{y}_i$.
	
	The univariate multiplicative seasonality case is given by
	\begin{align}
		\hat{y}_{t,~t + 1,~\dots,~t + m} &= \text{LSTM}(X_t) \odot \hat{l}_t \odot \hat{s}_{t,~t + 1,~\dots,~t + m} \label{eq:modified1}\\
		x_i &= \frac{y_i}{\hat{l}_i \hat{s}_i}~\textnormal{for } i = t + 1,~\dots,~t + m,
		\label{eq:x1}
	\end{align}
	where the LSTM takes as input a vector of de-trended and de-seasonalized observations of which a scalar element $x_i$ is computed using Equation~\eqref{eq:x1}.
	
	Finally, the multivariate multiplicative case is thus expressed as
	\begin{align}
		\hat{\boldsymbol{y}}_{t, t + 1 \dots t + m} &= \text{LSTM}(\boldsymbol{X}_t) \odot \hat{\boldsymbol{l}}_t \odot \hat{\boldsymbol{s}}_{t, t + 1 \dots t + m} \label{eq:mmodified1}\\
		\boldsymbol{x}_i &= \boldsymbol{y}_i \odot \left( \hat{\boldsymbol{l}}_i \odot \hat{\boldsymbol{s}}_i \right)^{-1}~\text{for } i = t + 1,~\dots,~t + m, \label{eq:mx1}
	\end{align}
	where the LSTM takes as input a size $k$ x $m$ matrix $\boldsymbol{X}_t$ of de-trended, de-seasonalized observations, where each vector component $\boldsymbol{x}_i$ is computed via Equation~\eqref{eq:mx1}.
	
	\subsection{Deep Learning Layer}
	\label{subsec:nnlayer}
	
	Our neural network data flow is presented in Figure~\ref{fig:meslstm_01}. We input a matrix composed of $k$ vectors each of size $m$. After preprocessing, we input the matrix to our LSTM layer. The LSTM layer output (composed of $j$ predictands each sized $n$) feeds into a dense layer, then postprocessing is conducted to produce the model's final output. LSTMs employ gated connections, are good at modeling latent representations with temporal dependency, and as a result, are superior to vanilla RNNs~\cite{Hoch97}. The deep learning and exponential smoothing layers are optimized consecutively, as depicted in Figure~\ref{fig:meslstm_01}.
	
	\begin{figure}[h]
		\centering 
		\includegraphics[width=0.85\columnwidth]{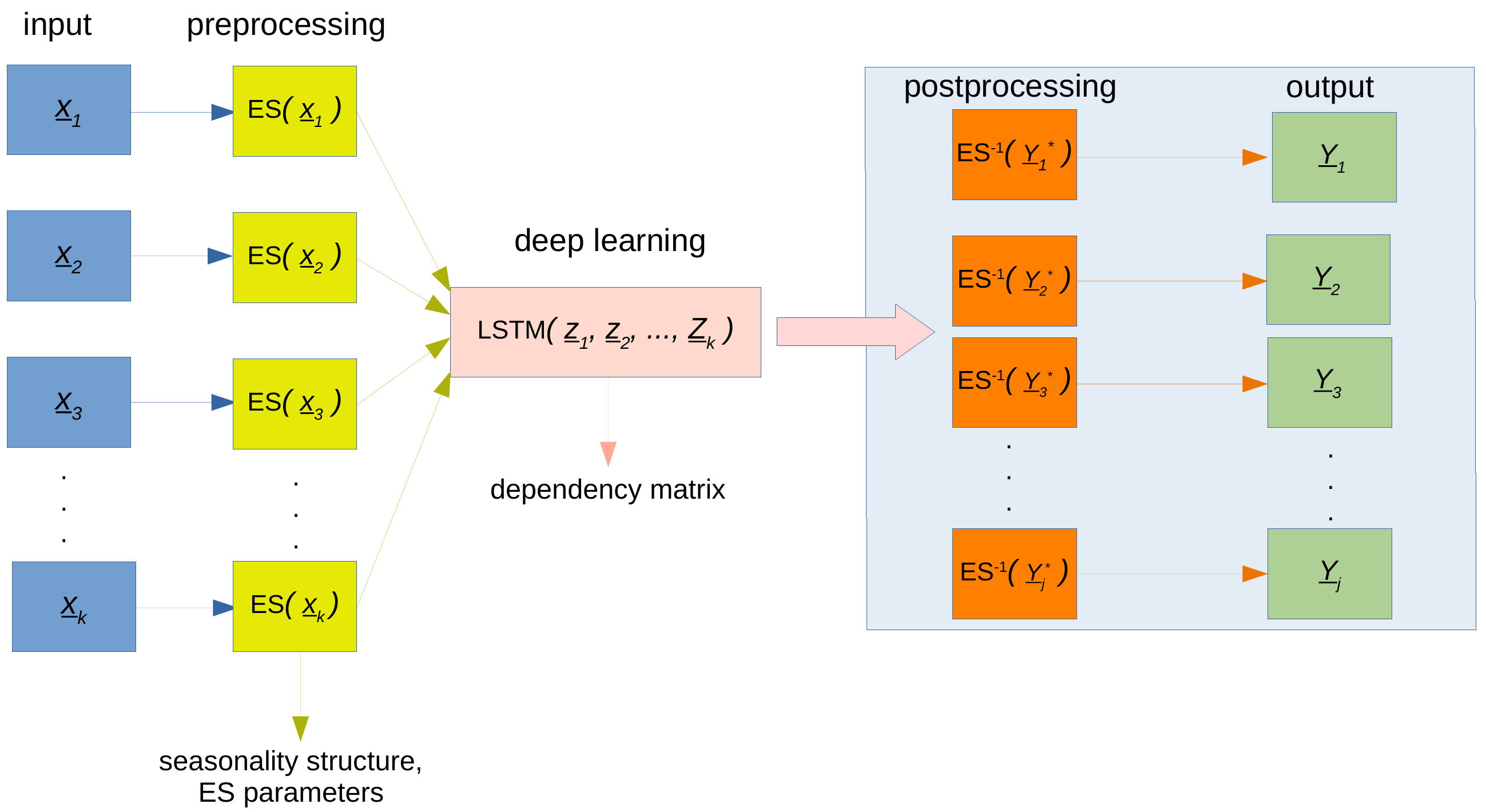}
		\caption{MES-LSTM Data Flowchart.\label{fig:meslstm_01}}
	\end{figure}
	
	In order to quantify the uncertainty associated with our point forecasts, we modify the above architecture as follows. Instead of the dense layer that back-propagates and learns scalar sets of weights and biases, we employ a densely-connected layer class with Flipout estimator~\cite{Wen18} that learns a \emph{distribution} of weights and biases. In this variational, probabilistic setting, traditional back-propagation is replaced by Bayes-by-backprop~\cite{Blund15}.
	
	Flipout~\cite{Wen18}, through Monte Carlo simulation, approximates a distribution of weights by integrating over the kernel and bias. By assuming the kernel and biases are drawn from distributions, the dense-flipout is able to implement the Bayesian Variational inference dense-flipout layer, analogous to the above (traditional) dense layer . Using samples from the kernel and bias posteriors, this dense-flipout layer is able to run stochastic forward passes. Another difference between this stochastic process and the analogous densely connected layer, is we use variational inferencing by minimizing the KL-divergence~\cite{Joy11}. The uncertainty quantification is implemented using Tensorflow Probability~\cite{Dil17}.
	
	In effect, we can now forecast using a sample from the distribution of weights and biases. Applying Monte-Carlo simulation, each time we forecast, we iteratively compile a distribution of forecasts. We then use the appropriate percentiles to extract the desired prediction intervals from the forecast distribution. So, to compute the $(1 - \alpha) \times 100 \%$ prediction interval, we use the percentiles at 
	\begin{align}
		\left(\dfrac{\alpha}{2}, \,\left(1 - \alpha\right) + \left( \dfrac{\alpha}{2}\right) \right).
	\end{align}
	
	The data flow process for quantifying uncertainty is not dissimilar to Figure~\ref{fig:meslstm_01}. The only difference is that in the deep learning layer, we now produce probabilistic forecasts instead of deterministic ones.
	
	The variational inference approach described above is similar to two other notable prediction interval construction techniques, i.e. Monte Carlo dropout~\cite{Gal16b} and the quantile bootstrap~\cite{Dav13} also known as the reverse percentile bootstrap~\cite{Hest14}. The technique used here, Flipout, is similar to dropout and bootstrapping as they all use quantiles extracted from synthetic distributions to produce the final intervals. \citet{Mat2} offer examples where dropout and bootstrapping have been applied and compared \cite{Mat2}.
	
	The key differences in all three methods is how their respective distributions are constructed, and their relative uncertainty quantification skill. See for instance the work of \citet{Wen18}, where Flipout has been shown to outperform dropout. In addition, a fortuitous side-effect of the variational inference (and dropout methods) is a reduction in epistemic uncertainty, which in turn regularizes the network and addresses any concerns that may arise from the possibility of a lack of sufficient training data, and issues associated with overfitting.
	
	To summarize, we input a matrix (sized number of training observations by number of predictors) to the model. The preprocessing layer computes all the exponential smoothing parameters as outlined in Subsection~\ref{subsec:prelayer} above. We then feed the output directly to the deep learning layer, which discerns the inherent dependency within the covariates. The first step in our two-step methodology is inspired by the works of Harvey~\cite{Hav86}, who, through simplifications, enables forecasting related time series through the use of univariate smoothing models. We find this two-step methodology for MES yields similar results to using the one-step formulation given by Pfeffermann~\cite{Pfe83}, with our method enjoying the added benefit of reduction in computational cost. The output from the LSTM is then fed directly to the dense layer or dense-flipout layer in the case of point forecasts and prediction interval construction, respectively. Finally, seasonality and trend estimates are added back to these intermediate outputs. We now have the final model forecasts and prediction intervals, respectively.
	
	Figure~\ref{fig:meslstm_02} illustrates our model architecture zoomed in to the deep learning layer. The input shape is $(k,\,m)$ where we have $k$ covariates each with $m$ observations in the training data. The network's intermediate output (before postprocessing) is generated by feeding the output of the LSTM through to a dense layer with a ReLU activation function. The size of the LSTM, $S$, is deduced empirically using the technique described in Section~\ref{sub:hyperparams} below. The size of the dense layer (or dense-flipout layer for uncertainty quantification) and correspondingly the output layer is determined by the number of predictands $j$. The output in the schematic then goes through postprocessing to meet our required format. It is re-trended and re-seasonlized (using the parameter estimates from the exponential smoothing equations) to arrive at the format presented by the ground truth data.
	
	\begin{figure}[htbp]
		\centering 
		\includegraphics[width=0.7\columnwidth]{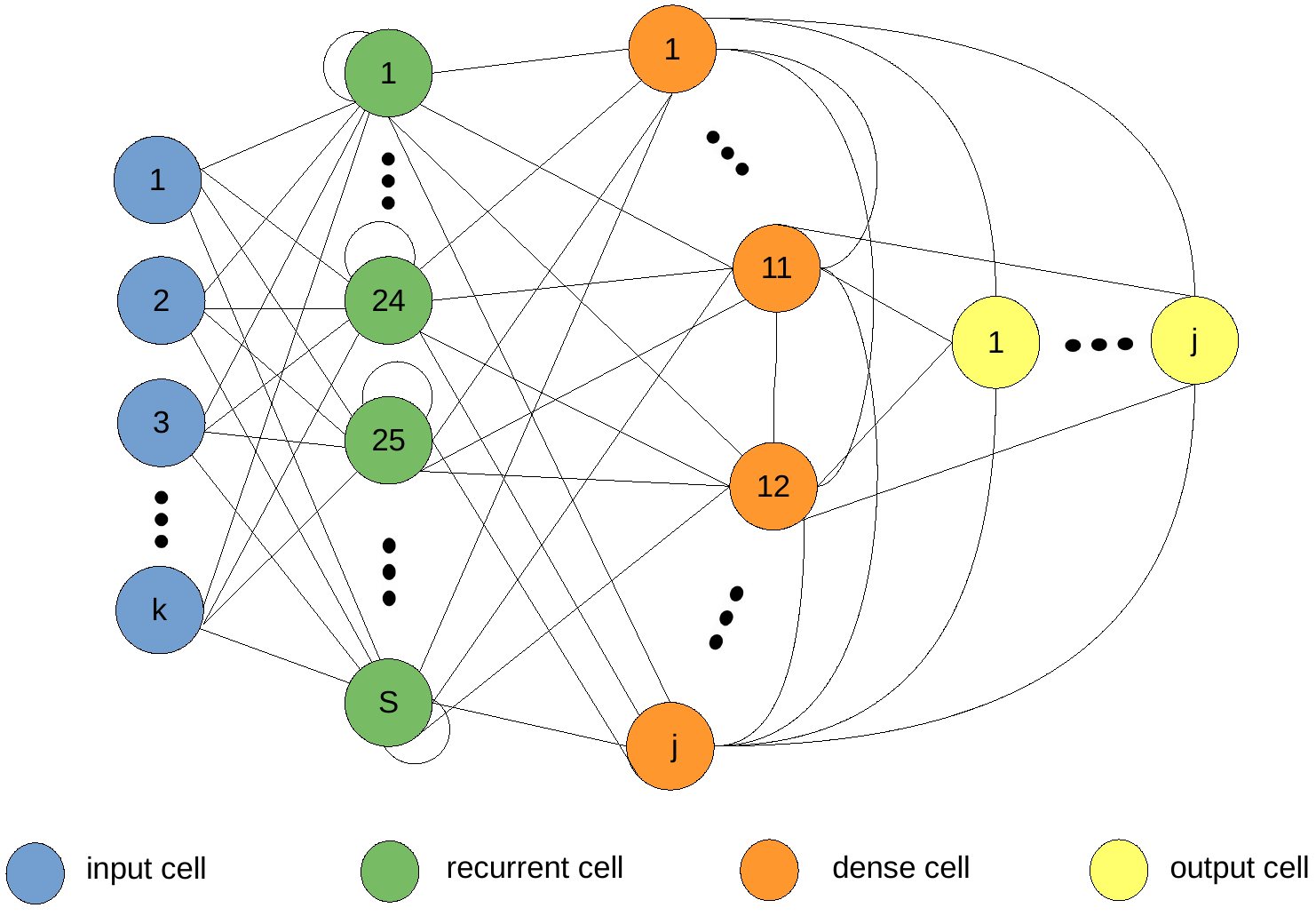}
		\caption{MES-LSTM Architecture Diagram.\label{fig:meslstm_02}}
	\end{figure}
	
	\subsection{Hyperparameter Tuning}
	\label{sub:hyperparams}
	
	Note from our formulation given in Equations~\eqref{eq:mmodified} and~\eqref{eq:mx} (multivariate model with additive seasonality), and Equations~\eqref{eq:mmodified1} and~\eqref{eq:mx1} (multivariate model with multiplicative seasonality), we do not need to compute the estimates for the trend component $\hat{\boldsymbol{b}}_t$ and this significantly reduces the hyper-parameter search space of the preprocessing layer. The initial estimates of the coefficients for level and seasonality are deduced by calculating primer estimates following the classical exponential smoothing Equations~\eqref{eq:level} and~\eqref{eq:season}, as well as Equations~\eqref{eq:level1} and~\eqref{eq:season1} for the respective additive and multiplicative seasonality cases. The level is initialized as the in-sample average, and the initial seasonality is the deviations between the level plus trend fit and the first week's observations. If we have $k$ attributes in our dataset, the model computes and stores $(k + 1) * (2 + P)$ exponential smoothing parameters, where $P$ indicates seasonality length.
	
	For the deep learning layer, we use a small model with relatively few parameters. The reasons for this model configuration are two-fold. First, we have a relatively small amount of data for training, and secondly, a large over-parameterized network might overfit and not generalize well to the test data~\cite{Lev16}.
	
	We iterate over a subset of LSTM sizes and training epochs and select the model configuration with the best forecast accuracy for our validation data and fewest parameters to fit. We optimize the batch size and number of samples the rolling window looks into the past to forecast the next example. We summarize our hyperparameter search space for the deep learning layer in Table~\ref{tab:hps}. We inspect the best five configurations (w.r.t. the error metrics detailed in the section that follows) in terms of forecasting all of the predictands. We then select the model configuration that is most parsimonious in order to ensure our model best mitigates overfitting over multiple runs. According to this methodology, the best configuration is given by LSTM of size 50, trained over 25 epochs, using batches sized 16, with an input window of 14 days.
	
	\begin{specialtable}[h!t]
		\centering
		\caption{Configuration Grid for Deep Learning Layer Hyperparameter Search Space.}
		\label{tab:hps}
		\begin{tabular}{lr}
			\toprule
			\multicolumn{1}{l}{\textbf{Hyperparameter}} & \multicolumn{1}{r}{\textbf{Search Space}} \\
			\bottomrule\toprule
			LSTM size    & 50, 55, 60, $\dots$, 150 \\
			epochs       & 15, 20, 25, $\dots$, 75  \\
			batch size   & 8, 16, 24, $\dots$, 64 \\
			input window & 7, 14, 21\\ \bottomrule
		\end{tabular}
	\end{specialtable}
	
	Our train-validation-test data split is 75-15-10. The reason for this is we forecast in the short-to-medium term, so 10\% test data is sufficient to evaluate our model's performance.
	
	The order of our VARMAX model, $\left( p, \, q \right)$, is deduced by conducting a grid search for optimization in much the same way as above for MES-LSTM. 
	We set the maximum iterations for maximum likelihood search to 200 (four times the default in Python) to ensure convergence. The trend component is deduced through a grid search varying the trend polynomial from constant, linear, quadratic, to cubic. We auto-regress the predictors and the predictands are declared as exogenous.
	
	The model hyperparameter optimization process for SARIMAX is exactly the same as for VARMAX. In both cases we perform the grid search and choose the model with the smallest Akaike information criteria (AIC,~\cite{Akaike73}). For simplicity, we do not enforce invertibility on the moving average polynomials. This also ensures more of the models are estimable. In cases where there are several suitable model configurations (equal AIC values) we choose the model that is most parsimonious (smallest product of $p$ and $q$).
	
	The MLR is fit using Ordinary Least Squares (OLS,~\cite{Mat2005}). We plug in the optimal hyperparameters for LSTM into the DeepAR, with additional searches for the dropout rate (0.1, 0.15, 0.2), and likelihood (noise model for probabilistic forecasts, so we can discern uncertainty) between Gaussian and student-T. The LGB model was tuned considering tree maximum depth (capped at 5), maximum leaves (capped at 30), number of estimators (capped at 125), and the learning rate (same search space as DeepAR).
	
	All the models are trained on a cluster of Intel Core i7-9750H processors each with an Nvidia GeForce GTX 1650 GPU and 4GB RAM.
	
	\subsection{Metrics}
	\label{subsec:metrics}
	For determining the most suitable seasonality structure in the preprocessing layer, we use the Sum of Squared Errors (SSE). For training in the deep learning layer, we use Mean Absolute Error loss (MAE) with Adam efficient optimizer~\cite{Die15}. For performance evaluation of the point forecasts, we employ the symmetric Mean Absolute Percent Error (sMAPE, \cite{Makri00}) and Root Mean Squared Error (RMSE),
	\begin{align}
		\text{sMAPE} &= \dfrac{2}{m}\sum_{t = 1}^{m}\dfrac{\left|y_t - \hat{y}_t \right|}{\left| y_t \right| + | \hat{y}_t |}\\
		\text{RMSE} &= \sqrt{\dfrac{1}{m}\sum_{t = 1}^{m}\left( y_t - \hat{y}_t \right)^2},
	\end{align}
	where $y_t$ is the post-sample value of the predictand at point $t$, $\hat{y}_t$ the estimated forecast and $m$ is again the forecast horizon.
	The former is in line with the performance metrics from the M4 Competition, and is used for continuity as results published by Smyl~\cite{Smyl20} also used this metric. Further motivation for our choice is sMAPE being a median-based error criterion, it is also useful in instances where there are large amounts of outliers in the data. Lastly, sMAPE is symmetric on an absolute scale~\cite{Koe2001}.
	
	The latter is useful as it gives error in the same units as the forecast variable itself, thus easily interpretable. The other reason we employ RMSE as a complement is although symmetric on an absolute scale, sMAPE has been shown to penalize large positive errors more than negative ones on a percentage scale~\cite{Good99}. Both metrics are mean absolute differences between forecast and actual values. The key difference is how sMAPE is normalized.
	
	For evaluating the prediction intervals we use the Mean Interval Score (MIS, \cite{Gneit07}), which is averaged over all out-of-sample observations,
	\begin{align}
		\label{eq:MIS}
		\text{MIS$_{\alpha}$} &= \dfrac{1}{m}\sum_{t = 1}^{m}\left( U_t - L_t \right) + \dfrac{2}{\alpha} \left( L_t - y_t \right) \mathds{1}_{\left\lbrace y_t < L_t \right\rbrace} + \dfrac{2}{\alpha} \left( y_t - U_t \right) \mathds{1}_{\left\lbrace y_t > U_t \right\rbrace},
	\end{align}
	where $U_t$ $\left( L_t \right)$ is the upper (lower) bound of the prediction interval at time $t$, $\alpha$ is the significance level and $\mathds{1}$ is the indicator function.
	
	The MIS$_{\alpha}$ adds a penalty at the points where future observations are outside the specified bounds $\left[ L_t, U_t\right]$. The width of the interval at $t$ is added to said penalty, if any. As such, the MIS$_{\alpha}$ also penalizes wide intervals. Finally, this sum at the individual out-of-sample points is averaged.
	
	As a supplementary metric for evaluating the performance of the prediction intervals, we employ the Coverage Score (CS$_{\alpha}$) which indicates the percentage of observations that fall within the prediction interval,
	\begin{align}
		\text{CS$_{\alpha}$} &= \dfrac{1}{m}\sum_{t = 1}^{m} \mathds{1}_{\left\lbrace U_t \leq y_t \leq L_t \right\rbrace}.
	\end{align}
	
	With MIS$_{\alpha}$ the subscript $\alpha$ denotes explicit dependence on the significance level, whereas in the case of CS$_{\alpha}$ the dependence is implicit.
	
	\subsection{Benchmarks}
	\label{subsec:bench}
	
	For benchmarking MES-LSTM, the statistical methods used are Multiple Linear Regression (MLR, \cite{Mat2005}), Vector Autoregression Moving-Average with Exogenous Regressors (VARMAX, \cite{Han88})
	and Seasonal Autoregressive Integrated Moving-Average with Exogenous Regressors (SARIMAX, \cite{Aru16}). For deep learning benchmarks we use a vanilla LSTM (without direct incorporation of a preprocessing layer as described in Subsection~\ref{subsec:prelayer},) and Light Gradient Boosting Machine (LGB, \cite{Ke17}).
	
	The choice of the pure deep learning benchmarks/baselines models is based on, for one, the architecture of our proposed model. Because we have a statistical-and-LSTM hybrid model, it makes sense to focus on LSTM w.r.t. pure deep learning techniques. Secondly, the choice of LGB (and LSTM) is motivated by the findings of the M5 Competitions. Almost all of the top 50 submissions for the M5 Accuracy~\cite{m5a} and Uncertainty~\cite{m5u} competitions use a variation of LGB. In particular, considering the top five: submissions ranked 1, 2, 4, and 5 from the Accuracy competition~\cite{m5a} incorporate LGB into their model, and the submission ranked third uses a deep autoregressive recurrent neural network (DeepAR,~\cite{Sali20}), which is built on multiple LSTMs. From the top five submissions in the M5 Uncertainty competition~\cite{m5u}, 1, 2, 3, and 5 incorporate LGB, while the submission ranked fourth incorporates LSTM.

	
	\section{Datasets}
	\label{sec:data}
	We use the Our World in Data (OWID) COVID-19 dataset \cite{Mat21, Has20}. This dataset is aggregated from various sources, and includes historical data on the pandemic up to the date of publication, updated daily. A summary of the data and the aggregated data sources is shown in Table~\ref{tab:data_01}. It is important to note that even though the actual databases are updated at daily, weekly, and other time frequencies, the aggregated datasets used in our study are all presented at a daily temporal resolution. This means no frequency harmonization is required.
	
	\begin{specialtable}[H]
		\centering
		\caption{OWID COVID-19 Dataset Summary.}
		\label{tab:data_01}
		\resizebox{\columnwidth}{!}{%
			\begin{tabular}{lllr}
				\toprule
				\multicolumn{1}{l}{\textbf{Metrics}} & \multicolumn{1}{c}{\textbf{Source}} & \multicolumn{1}{l}{\textbf{Updated}} & \multicolumn{1}{r}{\textbf{Countries}} \\
				\bottomrule\toprule
				Vaccinations                & Official data collated by the Our World in Data team      & Daily  & 217 \\
				Tests \& positivity         & Official data collated by the Our World in Data team      & Weekly & 136 \\
				Hospital \& ICU             & Official data collated by the Our World in Data team      & Weekly & 34  \\
				Confirmed cases             & JHU CSSE COVID-19 Data                                    & Daily  & 194 \\
				Confirmed deaths            & JHU CSSE COVID-19 Data                                    & Daily  & 194 \\
				Reproduction rate           & Arroyo-Marioli F, Bullano F, Kucinskas S, Rondón-Moreno C & Daily  & 184 \\
				Policy responses            & Oxford COVID-19 Government Response Tracker               & Daily  & 186 \\
				Other variables of interest & International organizations (UN, World Bank, OECD, IHME…) & Fixed  & 240 \\
				\bottomrule
		\end{tabular}}
	\end{specialtable}
	
	The variables represent data related to confirmed cases, deaths, hospitalizations, and testing, as well as 56 other variables. We only use a subset of the available attributes, as detailed in the feature list shown in Table~\ref{tab:data_02}.
	
\end{paracol}
\begin{specialtable}[h!t]
	\centering
	\caption{OWID COVID-19 Dataset Feature List.}
	\label{tab:data_02}
	\resizebox{0.975\columnwidth}{!}{
		\begin{tabular}{ll}
			\toprule
			\multicolumn{1}{l}{\textbf{Variable}}  & \multicolumn{1}{c}{\textbf{Description}} \\
			\bottomrule\toprule
			total cases &
			Total confirmed cases of COVID-19 \\
			new cases &
			New confirmed cases of COVID-19 \\
			total cases per million &
			Total confirmed cases of COVID-19 per 1,000,000 people \\
			new cases per million &
			New confirmed cases of COVID-19 per 1,000,000 people \\
			total deaths &
			Total deaths attributed to COVID-19 \\
			new deaths &
			New deaths attributed to COVID-19 \\
			total deaths per million &
			Total deaths attributed to COVID-19 per 1,000,000 people \\
			new deaths per million &
			New deaths attributed to COVID-19 per 1,000,000 people \\
			icu patients &
			Number of COVID-19 patients in intensive care units (ICUs) on a given day \\
			icu patients per million &
			Number of COVID-19 patients in ICUs on a given day per 1,000,000 people\\
			hosp patients &
			Number of COVID-19 patients in hospital on a given day \\
			weekly icu admissions &
			Number of COVID-19 patients newly admitted to ICUs in a given week\\
			weekly icu admissions per million &
			Number of COVID-19 patients newly admitted to ICUs in a given week per 1,000,000 people\\
			weekly hosp admissions &
			Number of COVID-19 patients newly admitted to hospitals in a given week \\
			weekly hosp admissions per million &
			Number of COVID-19 patients newly admitted to hospitals in a given week per 1,000,000 people\\
			stringency index &
			Government Response Stringency Index: composite measure based on 9 response indicators\\
			reproduction rate &
			Real-time estimate of the effective reproduction rate (R) of COVID-19 \\
			total tests &
			Total tests for COVID-19 \\
			new tests &
			New tests for COVID-19 (only calculated for consecutive days) \\
			positive rate &
			Share of COVID-19 tests that are positive, rolling 7-day average (inverse of tests per case) \\
			tests per case &
			Tests conducted per new confirmed case of COVID-19, rolling 7-day average (inverse of positive rate) \\
			total vaccinations &
			Total number of COVID-19 vaccination doses administered \\
			people vaccinated &
			Total number of people who received at least one vaccine dose \\
			people fully vaccinated &
			Total number of people who received all doses \\
			new vaccinations &
			New COVID-19 vaccination doses administered (only calculated for consecutive days) \\
			total vaccinations per hundred &
			Total number of COVID-19 vaccination doses administered per 100 people \\
			people vaccinated per hundred &
			Total number of people who received at least one vaccine dose per 100 people \\
			people fully vaccinated per hundred &
			Total number of people who received all doses prescribed by the vaccination protocol per 100 people \\
			location &
			Geographical location \\
			date &
			Date of observation \\
			population &
			Population in 2020 \\
			population density &
			Number of people divided by land area, measured in square kilometers \\
			median age &
			Median age of the population, UN projection for 2020 \\
			aged 65 older &
			Share of the population that is 65 years and older, most recent year available \\
			aged 70 older &
			Share of the population that is 70 years and older in 2015 \\
			gdp per capita &
			Gross domestic product at purchasing power parity \\
			extreme poverty &
			Share of the population living in extreme poverty\\
			cardiovasc death rate &
			Death rate from cardiovascular disease in 2017 \\
			diabetes prevalence &
			Diabetes prevalence (\% of population aged 20 to 79) in 2017 \\
			female smokers &
			Share of women who smoke, most recent year available \\
			male smokers &
			Share of men who smoke, most recent year available \\
			handwashing facilities &
			Share of the population with basic handwashing facilities on premises \\
			hospital beds per thousand &
			Hospital beds per 1,000 people, most recent year available since 2010 \\
			life expectancy &
			Life expectancy at birth in 2019 \\
			human development index &
			Composite average achievement in  ($i$) a long, healthy life ($ii$) knowledge ($iii$) standard of living \\
			excess mortality &
			Excess mortality P-scores for all ages\\
			\bottomrule
		\end{tabular}
	}
\end{specialtable}
\begin{paracol}{2}
	\switchcolumn
	
	We choose this aggregated dataset over other available COVID-19 datasets because it does to some extent mitigate the concerns raised by~\citet{Roh21}, for example. As highlighted in Section~\ref{sub:litreview}, some exogenous factors not directly related to COVID-19 may be useful in the modeling process. For instance, the OWID COVID-19 dataset includes covariates such as \textit{human\_development\_index}, \textit{extreme\_poverty}, and \textit{handwashing\_facilities}.
	
	The variables that we have omitted from our study are duplicates where numerical data has been smoothed, e.g. \textit{new\_cases\_smoothed.} The smoothed attributes are removed for two reasons: OWID offers no insights on how the data is smoothed, and we avoid any duplication as our model also conducts preprocessing.
	
	In particular, we use our model to forecast and quantify the associated prediction uncertainty for two attributes of interest: \textit{total cases} and \textit{total deaths}. The predictions and uncertainty quantification can be conducted for a single country, or easily across a multitude of countries or even averaged over a specific region. Regional or multi-country inferencing can assist policy-makers to make tough decisions like restricting inter-provincial/state travel or closing national borders completely.
	
	Below we present results for the Southern African Development Community (SADC). SADC is a regional economic community comprising 16 member states: Angola, Botswana, Comoros, Democratic Republic of Congo (DRC), Eswatini, Lesotho, Madagascar, Malawi, Mauritius, Mozambique, Namibia, Seychelles, South Africa,  Tanzania, Zambia and Zimbabwe.
	
	\section{Results and Discussion}
	\label{sec:results}
	In this section, we detail the results for multivariate point forecasts as well as the prediction intervals for our predictands \textit{total cases} and \textit{total deaths}. The results are presented for the analysis conducted for the SADC region. In order to mitigate the stochastic nature of LSTMs, the experiments are repeated 35 times. The results presented are the aggregates of all the repeated independent trials. 
	
	
	\subsection{Forecast Performance for SADC}
	The results for point forecasts for SADC are tabulated in Table~\ref{tab:sadc_cases} and~\ref{tab:sadc_deaths}, for  each of our predictands. We note MES-LSTM outperforms the benchmarks for all the nations in SADC, except sMAPE for \textit{total cases} in South Africa. The best performer in each instance is highlighted for ease of interpretation. Interestingly, overall, the worst forecast performance from MES-LSTM is in South Africa and this could be a result of the nation presenting the most accurate data. The other nations don't update their data as frequently, and the model learns easier as there isn't a lot of variability. Furthermore, the (less accurate) data from the other nations is closer to the linear assumptions of the statistical benchmark models, i.e. VARMAX, SARIMAX and MLR.
	
	\begin{specialtable}[h]
		\centering
		\caption{Forecast Accuracy Averaged Over All Trials for \textit{total cases} in the SADC Region.}
		\label{tab:sadc_cases}
		\resizebox{\columnwidth}{!}{%
			\begin{tabular}{lrr|rr|rr|rr|rr|rr|rr}
				\toprule
				&
				\multicolumn{2}{c}{\textbf{MES-LSTM}} &
				\multicolumn{2}{c}{\textbf{LSTM}} &
				\multicolumn{2}{c}{\textbf{LGB}} &

				\multicolumn{2}{c}{\textbf{DeepAR}} &
				\multicolumn{2}{c}{\textbf{VARMAX}} &
				\multicolumn{2}{c}{\textbf{SARIMAX}} &
				\multicolumn{2}{c}{\textbf{MLR}} \\ 
				\textbf{Country} &
				\multicolumn{1}{r}{\textbf{sMAPE}} &
				\multicolumn{1}{r}{\textbf{RMSE}} &
				\multicolumn{1}{|r}{\textbf{sMAPE}} &
				\multicolumn{1}{r}{\textbf{RMSE}} &
				\multicolumn{1}{|r}{\textbf{sMAPE}} &
				\multicolumn{1}{r}{\textbf{RMSE}} &
				\multicolumn{1}{|r}{\textbf{sMAPE}} &
				\multicolumn{1}{r}{\textbf{RMSE}} &
				\multicolumn{1}{|r}{\textbf{sMAPE}} &
				\multicolumn{1}{r}{\textbf{RMSE}} &
				\multicolumn{1}{|r}{\textbf{sMAPE}} &
				\multicolumn{1}{r}{\textbf{RMSE}} &
				\multicolumn{1}{|r}{\textbf{sMAPE}} &
				\multicolumn{1}{r}{\textbf{RMSE}} \\
				\bottomrule\toprule
\textbf{Angola} & \h0.7 & \h563.1 & 68.9 & 28023.4 & 5.6 & 28524.5 & 59.0 & 23811.6 & 76.4 & 35442.3 & 107.1 & 59732.6 & 77.6 & 35830.6 \\
\textbf{Botswana} & \h1.6 & \h5817.7 & 85.5 & 94414.1 & 5.8 & 32635.2 & 62.9 & 72983.0 & 99.5 & 125387.6 & 71.0 & 123392.6 & 114.6 & 137334.7 \\
\textbf{Comoros} & \h1.1 & \h52.5 & 16.5 & 623.3 & 5.3 & 32005.1 & 50.1 & 1443.4 & 6.2 & 313.8 & 23.6 & 1058.6 & 17.6 & 704.2 \\
\textbf{DRC} & \h0.8 & \h468.3 & 72.6 & 25987.3 & 5.3 & 29933.9 & 41.9 & 17302.5 & 12.8 & 7047.8 & 27.6 & 15527.6 & 83.9 & 34102.8 \\
\textbf{Eswatini} & \h1.3 & \h617.7 & 59.1 & 18348.9 & 5.7 & 31572.0 & 61.2 & 17619.4 & 63.6 & 23417.0 & 37.9 & 16687.6 & 110.6 & 33055.4 \\
\textbf{Lesotho} & \h0.7 & \h167.2 & 47.2 & 7324.4 & 5.5 & 28662.4 & 33.9 & 5484.3 & 35.2 & 6478.7 & 137.8 & 24942.5 & 42.4 & 7603.3 \\
\textbf{Madagascar} & \h1.3 & \h636.6 & 35.1 & 12019.8 & 5.5 & 29387.4 & 45.6 & 13719.5 & 11.1 & 5495.1 & 8.8 & 3889.1 & 22.3 & 9014.0 \\
\textbf{Malawi} & \h1.3 & \h817.5 & 22.0 & 11967.0 & 5.9 & 30163.2 & 14.1 & 8183.4 & 10.4 & 7629.0 & 9.7 & 5956.9 & 42.2 & 23504.7 \\
\textbf{Mauritius} & \h3.8 & \h2056.8 & 99.8 & 9891.6 & 5.8 & 29677.4 & 91.1 & 8643.8 & 179.5 & 17135.7 & 242.9 & 32875.6 & 171.3 & 16742.4 \\
\textbf{Mozambique} & \h1.1 & \h1685.0 & 5.0 & 7457.6 & 5.2 & 29234.1 & 3.0 & 4438.9 & 2.5 & 3946.2 & 90.5 & 125768.7 & 10.3 & 15089.2 \\
\textbf{Namibia} & \h1.1 & \h1508.8 & 5.2 & 7516.3 & 5.6 & 26149.9 & 2.3 & 2969.0 & 12.0 & 17579.1 & 25.3 & 32551.7 & 7.7 & 10010.1 \\
\textbf{Seychelles} & \h0.9 & \h266.7 & 59.9 & 8908.6 & 5.4 & 27596.0 & 62.1 & 8596.8 & 11.4 & 2428.1 & 47.4 & 10115.2 & 99.0 & 14797.0 \\
\textbf{South Africa} & 5.0 & \h26979.2 & 4.8 & 150416.4 & 5.6 & 30420.9 & 7.8 & 212612.0 & 2.8 & 87376.0 & 8.2 & 265998.0 & \h4.0 & 118139.5 \\
\textbf{Tanzania} & \h0.3 & \h134.6 & 116.3 & 15445.1 & 6.0 & 30900.7 & 96.5 & 12839.8 & 184.1 & 25060.9 & 636.9 & 87188.8 & 194.9 & 25818.8 \\
\textbf{Zambia} & \h1.2 & \h2499.0 & 6.4 & 15567.7 & 5.8 & 28686.9 & 2.0 & 4333.4 & 7.3 & 16766.9 & 72.1 & 143490.3 & 2.1 & 6120.9 \\
\textbf{Zimbabwe} & \h1.1 & \h1518.5 & 8.6 & 10652.2 & 5.4 & 32284.2 & 5.7 & 7251.2 & 8.4 & 12296.7 & 6.4 & 9733.6 & 2.6 & 4164.0 \\
				\bottomrule
		\end{tabular}}
	\end{specialtable}
	
	\begin{specialtable}[h]
		\centering
		\caption{Forecast Accuracy Averaged Over All Trials for \textit{total deaths} in the SADC Region.}
		\label{tab:sadc_deaths}
		\resizebox{\columnwidth}{!}{%
			\begin{tabular}{lrr|rr|rr|rr|rr|rr|rr}
				\toprule
				&
				\multicolumn{2}{c}{\textbf{MES-LSTM}} &
				\multicolumn{2}{c}{\textbf{LSTM}} &
				\multicolumn{2}{c}{\textbf{LGB}} &

				\multicolumn{2}{c}{\textbf{DeepAR}} &
				\multicolumn{2}{c}{\textbf{VARMAX}} &
				\multicolumn{2}{c}{\textbf{SARIMAX}} &
				\multicolumn{2}{c}{\textbf{MLR}} \\ 
				\textbf{Country} &
				\multicolumn{1}{r}{\textbf{sMAPE}} &
				\multicolumn{1}{r}{\textbf{RMSE}} &
				\multicolumn{1}{|r}{\textbf{sMAPE}} &
				\multicolumn{1}{r}{\textbf{RMSE}} &
				\multicolumn{1}{|r}{\textbf{sMAPE}} &
				\multicolumn{1}{r}{\textbf{RMSE}} &
				\multicolumn{1}{|r}{\textbf{sMAPE}} &
				\multicolumn{1}{r}{\textbf{RMSE}} &
				\multicolumn{1}{|r}{\textbf{sMAPE}} &
				\multicolumn{1}{r}{\textbf{RMSE}} &
				\multicolumn{1}{|r}{\textbf{sMAPE}} &
				\multicolumn{1}{r}{\textbf{RMSE}} &
				\multicolumn{1}{|r}{\textbf{sMAPE}} &
				\multicolumn{1}{r}{\textbf{RMSE}} \\
				\bottomrule\toprule
\textbf{Angola} & \h0.9 & \h19.8 & 74.6 & 783.5 & 6.6 & 2886.7 & 60.8 & 687.0 & 89.9 & 1055.7 & 71.6 & 1038.1 & 89.4 & 1052.1 \\
\textbf{Botswana} & \h1.0 & \h26.1 & 86.1 & 1203.9 & 6.7 & 2909.9 & 77.2 & 1128.8 & 97.6 & 1576.5 & 15.7 & 363.9 & 120.8 & 1806.0 \\
\textbf{Comoros} & \h1.5 & \h2.4 & 12.5 & 16.7 & 6.2 & 3069.4 & 72.4 & 66.8 & 15.6 & 27.3 & 13.5 & 20.6 & 10.9 & 16.0 \\
\textbf{DRC} & \h1.0 & \h12.0 & 67.2 & 470.1 & 6.1 & 3059.6 & 38.2 & 326.4 & 2.4 & 34.2 & 11.6 & 128.9 & 74.0 & 594.3 \\
\textbf{Eswatini} & \h1.1 & \h14.2 & 49.6 & 435.1 & 7.0 & 2594.2 & 58.8 & 490.0 & 46.3 & 525.6 & 35.3 & 403.1 & 95.1 & 801.0 \\
\textbf{Lesotho} & \h0.8 & \h5.4 & 47.2 & 222.9 & 6.9 & 2790.2 & 42.4 & 207.1 & 46.8 & 249.7 & 97.4 & 516.2 & 44.5 & 240.8 \\
\textbf{Madagascar} & \h1.2 & \h12.2 & 48.1 & 330.3 & 6.6 & 2820.6 & 71.5 & 431.7 & 4.2 & 44.6 & 2.0 & 20.4 & 41.7 & 335.9 \\
\textbf{Malawi} & \h1.2 & \h28.7 & 27.1 & 518.9 & 6.6 & 3072.6 & 20.4 & 416.1 & 14.5 & 349.6 & 14.2 & 314.9 & 46.2 & 958.7 \\
\textbf{Mauritius} & \h1.7 & \h39.9 & 103.0 & 109.6 & 6.9 & 3014.0 & 104.4 & 110.3 & 173.1 & 183.1 & 300.2 & 341.7 & 172.7 & 183.1 \\
\textbf{Mozambique} & \h1.1 & \h21.9 & 5.1 & 96.7 & 6.4 & 3269.7 & 4.5 & 85.6 & 4.9 & 98.7 & 9.2 & 180.1 & 15.6 & 281.4 \\
\textbf{Namibia} & \h1.1 & \h41.9 & 7.9 & 333.8 & 6.1 & 3200.6 & 4.9 & 174.9 & 19.6 & 848.9 & 17.2 & 601.2 & 29.3 & 913.9 \\
\textbf{Seychelles} & \h1.7 & \h7.0 & 71.9 & 54.0 & 7.1 & 3288.2 & 82.4 & 58.9 & 5.9 & 8.3 & 43.5 & 47.7 & 116.9 & 88.8 \\
\textbf{South Africa} & \h4.9 & \h446.2 & 7.8 & 7902.4 & 5.9 & 2773.4 & 10.4 & 8515.8 & 10.7 & 10806.6 & 16.1 & 15438.9 & 6.5 & 6276.2 \\
\textbf{Tanzania} & \h0.3 & \h3.8 & 114.7 & 425.6 & 6.7 & 3143.2 & 113.7 & 423.5 & 179.3 & 685.9 & 362.3 & 1379.1 & 193.0 & 712.9 \\
\textbf{Zambia} & \h1.3 & \h47.7 & 6.6 & 266.8 & 6.6 & 2806.5 & 4.3 & 174.2 & 7.5 & 301.4 & 1.9 & 87.8 & 11.7 & 420.1 \\
\textbf{Zimbabwe} & \h1.1 & \h53.4 & 6.5 & 298.2 & 6.3 & 3067.7 & 8.1 & 361.0 & 14.8 & 791.5 & 5.8 & 289.7 & 6.7 & 345.0 \\		\bottomrule
		\end{tabular}}
	\end{specialtable}
	
	In Figure~\ref{fig:bar_forecast_sadc} we average the forecast results across the entirety of the SADC region. We note MES-LSTM is the best aggregate performer. This figure also illustrates the importance of choosing multiple error metrics. In Figure~\ref{fig:smape_bar_sadc} SARIMAX for instance, seems to perform fairly poorly for both predictands, but this as a stand-alone interpretation would be inaccurate. From Figure~\ref{fig:rmse_bar_sadc} we note that in terms of regional skill for \textit{total deaths}, SARIMAX is actually competitive. Recall that RMSE gives the error in the same units as the original predictand. South Africa has a total population of 65 million, so any model that over- or under-forecasts \textit{total cases} by a few thousand cases could still be useful for planning and management of the pandemic outbreak in question.
	
	\begin{figure}[htbp]
		\begin{subfigure}{0.49\columnwidth}
			\includegraphics[width=\columnwidth]{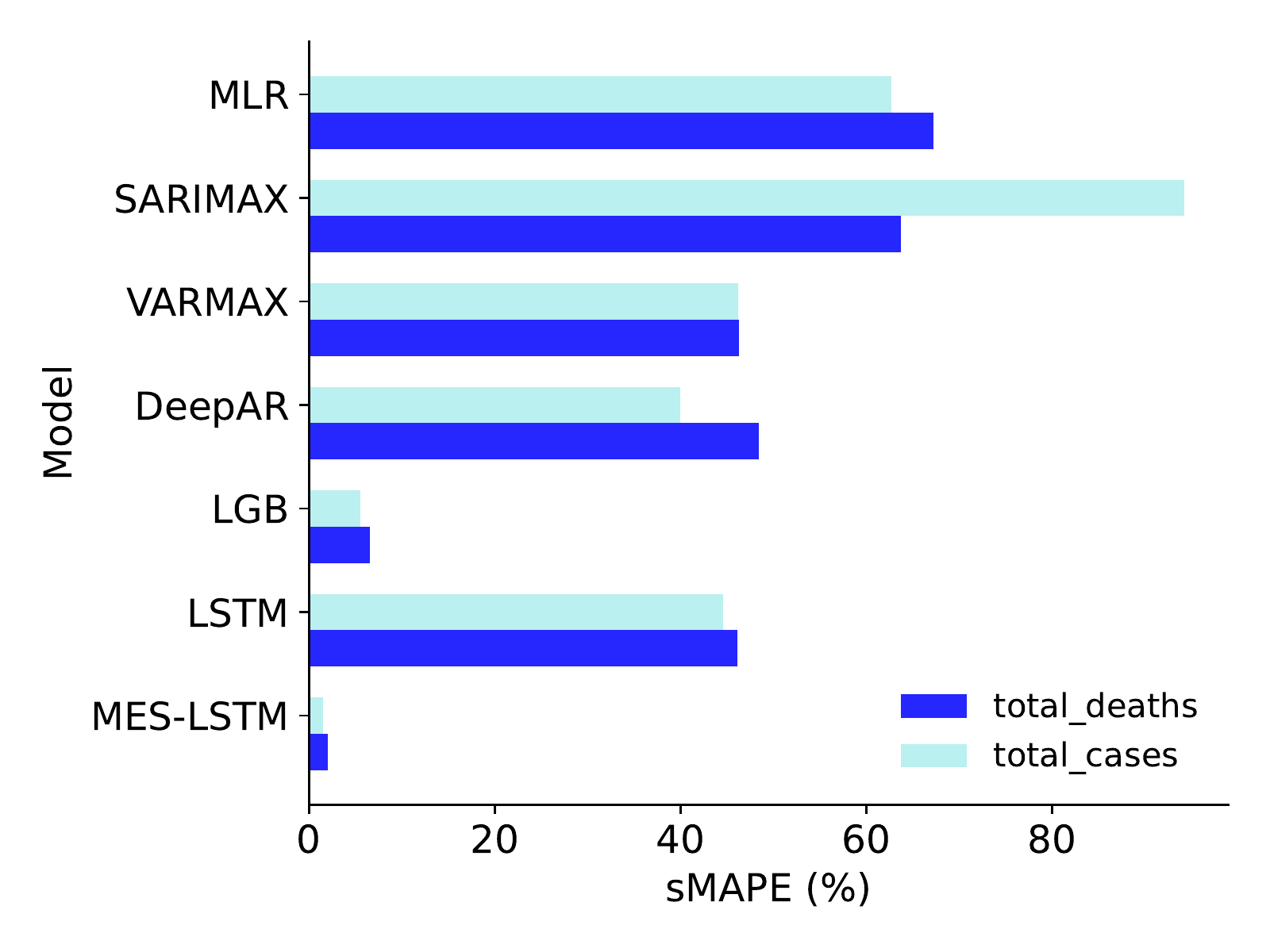}
			\caption{sMAPE}
			\label{fig:smape_bar_sadc}
		\end{subfigure}
		\begin{subfigure}{0.49\columnwidth}
			\includegraphics[width=\columnwidth]{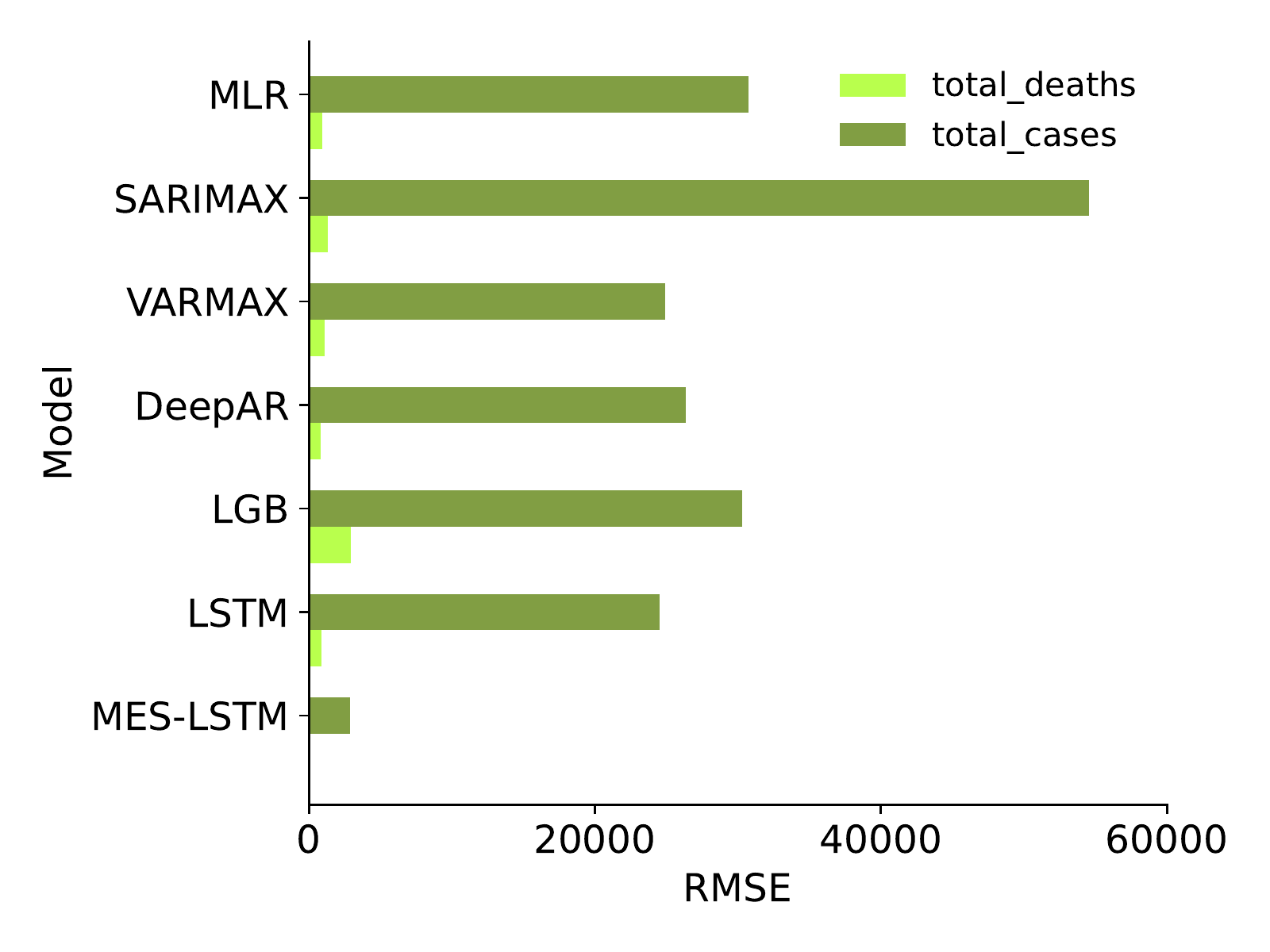}
			\caption{RMSE.}
			\label{fig:rmse_bar_sadc}
		\end{subfigure}
		\caption{Forecast Accuracy in the SADC Region.\label{fig:bar_forecast_sadc}}
		
	\end{figure}
	
	\subsection{Prediction Interval Performance for SADC}
	The results for the prediction interval are documented in Tables~\ref{tab:sadc_cases_pi_01} and~\ref{tab:sadc_deaths_pi_01}. The best performer or best tied performers are highlighted. We note the MES-LSTM MIS is superior to the other models for almost the entire SADC region.
	
	\begin{specialtable}[h]
		\centering
		\caption{Prediction Interval Accuracy Averaged Over All Trials for \textit{total cases} in SADC ($\alpha = 0.05$).}
		\label{tab:sadc_cases_pi_01}
		\resizebox{\columnwidth}{!}{%
		\begin{tabular}{lrr|rr|rr|rr|rr|rr|rr}
				\toprule
				&
				\multicolumn{2}{c}{\textbf{MES-LSTM}} &
				\multicolumn{2}{c}{\textbf{LSTM}} &
				\multicolumn{2}{c}{\textbf{LGB}} &

				\multicolumn{2}{c}{\textbf{DeepAR}} &
				\multicolumn{2}{c}{\textbf{VARMAX}} &
				\multicolumn{2}{c}{\textbf{SARIMAX}} &
				\multicolumn{2}{c}{\textbf{MLR}} \\ 
				\textbf{Country} &
				\multicolumn{1}{r}{\textbf{MIS}} &
				\multicolumn{1}{r}{\textbf{Coverage}} &
				\multicolumn{1}{|r}{\textbf{MIS}} &
				\multicolumn{1}{r}{\textbf{Coverage}} &
				\multicolumn{1}{|r}{\textbf{MIS}} &
				\multicolumn{1}{r}{\textbf{Coverage}} &
				\multicolumn{1}{|r}{\textbf{MIS}} &
				\multicolumn{1}{r}{\textbf{Coverage}} &
				\multicolumn{1}{|r}{\textbf{MIS}} &
				\multicolumn{1}{r}{\textbf{Coverage}} &
				\multicolumn{1}{|r}{\textbf{MIS}} &
				\multicolumn{1}{r}{\textbf{Coverage}} &
				\multicolumn{1}{|r}{\textbf{MIS}} &
				\multicolumn{1}{r}{\textbf{Coverage}} \\
				\bottomrule\toprule
\textbf{Angola} & \h1910.8 & \h95.2 & 116739.1 & 0.0 & 55663.7 & 81.6 & 54875.9 & 0.0 & 902714.3 & 0.0 & 814594.0 & 0.0 & 660826.0 & 0.0 \\
\textbf{Botswana} & \h11735.6 & \h85.2 & 378936.2 & 0.0 & 164794.4 & 77.6 & 171350.6 & 0.0 & 4572270.6 & 0.0 & 1203716.2 & 0.0 & 4259387.7 & 0.0 \\
\textbf{Comoros} & \h172.9 & 97.4 & 360.0 & 87.6 & 3908.4 & 84.4 & 5064.1 & 0.0 & 3085.7 & 40.9 & 1790.1 & 59.1 & 5461.2 & \h100.0 \\
\textbf{DRC} & \h2351.8 & \h97.6 & 102410.9 & 0.0 & 45538.0 & 65.2 & 68391.5 & 0.0 & 143127.3 & 0.0 & 137288.5 & 0.0 & 413486.5 & 0.0 \\
\textbf{Eswatini} & \h1837.3 & \h91.0 & 86314.2 & 0.0 & 44987.2 & 79.6 & 44252.9 & 0.0 & 518283.5 & 14.3 & 140160.1 & 0.0 & 683626.1 & 0.0 \\
\textbf{Lesotho} & \h750.4 & \h97.4 & 24269.2 & 0.4 & 20065.3 & 89.0 & 18693.3 & 0.0 & 218810.3 & 0.0 & 310359.2 & 0.0 & 24172.3 & 55.8 \\
\textbf{Madagascar} & \h2310.7 & 83.6 & 25720.5 & 16.9 & 30953.7 & 83.1 & 19934.0 & 56.3 & 139138.3 & 0.0 & 5270.5 & \h100.0 & 48987.3 & \h100.0 \\
\textbf{Malawi} & \h2503.3 & 88.0 & 27408.2 & 38.8 & 40635.9 & 78.9 & 31982.9 & 0.0 & 60588.0 & 70.2 & 33628.7 & \h100.0 & 303117.6 & 55.3 \\
\textbf{Mauritius} & \h2460.2 & \h90.1 & 44971.4 & 0.0 & 21957.8 & 77.6 & 43120.5 & 0.0 & 625931.1 & 0.0 & 441038.4 & 0.0 & 592480.4 & 0.0 \\
\textbf{Mozambique} & \h5249.2 & 94.2 & 17924.2 & 92.1 & 68002.5 & 86.0 & 3197.7 & 46.2 & 23040.1 & \h100.0 & 1364340.7 & 0.0 & 83378.3 & 22.9 \\
\textbf{Namibia} & \h4487.2 & \h93.1 & 18930.5 & 93.9 & 53947.4 & 79.7 & 1726.9 & 59.2 & 489847.0 & 0.0 & 28333.2 & 38.8 & 45068.1 & 67.3 \\
\textbf{Seychelles} & \h728.2 & \h93.1 & 32422.8 & 0.0 & 20364.2 & 73.7 & 17835.9 & 14.3 & 68568.1 & 0.0 & 103106.7 & 0.0 & 232264.3 & 0.0 \\
\textbf{South Africa} & \h193075.6 & 80.7 & 348860.7 & 97.7 & 1418713.2 & 86.6 & 239009.3 & \h100.0 & 740131.9 & \h100.0 & 1061162.0 & \h100.0 & 828447.2 & \h100.0 \\
\textbf{Tanzania} & \h943.6 & \h96.2 & 67427.5 & 0.0 & 32459.5 & 77.3 & 67350.0 & 0.0 & 990720.2 & 0.0 & 1279781.1 & 0.0 & 1018128.2 & 0.0 \\
\textbf{Zambia} & \h8972.0 & \h89.2 & 24151.6 & 96.4 & 89095.3 & 73.5 & 8316.6 & 16.6 & 489501.1 & 0.0 & 1509742.0 & 0.0 & 36167.6 & 95.8 \\
\textbf{Zimbabwe} & \h4770.4 & 87.7 & 23667.6 & 85.8 & 63299.2 & 77.9 & 2987.9 & \h100.0 & 367165.9 & 0.0 & 14904.8 & 95.8 & 16359.3 & \h100.0 \\
				\bottomrule
		\end{tabular}}
	\end{specialtable}
	
	\begin{specialtable}[H]
		\centering
		\caption{Prediction Interval Accuracy Averaged Over All Trials for \textit{total deaths} in SADC ($\alpha = 0.05$).}
		\label{tab:sadc_deaths_pi_01}
		\resizebox{\columnwidth}{!}{%
			\begin{tabular}{lrr|rr|rr|rr|rr|rr|rr}
				\toprule
				&
				\multicolumn{2}{c}{\textbf{MES-LSTM}} &
				\multicolumn{2}{c}{\textbf{LSTM}} &
				\multicolumn{2}{c}{\textbf{LGB}} &

				\multicolumn{2}{c}{\textbf{DeepAR}} &
				\multicolumn{2}{c}{\textbf{VARMAX}} &
				\multicolumn{2}{c}{\textbf{SARIMAX}} &
				\multicolumn{2}{c}{\textbf{MLR}} \\ 
				\textbf{Country} &
				\multicolumn{1}{r}{\textbf{MIS}} &
				\multicolumn{1}{r}{\textbf{Coverage}} &
				\multicolumn{1}{|r}{\textbf{MIS}} &
				\multicolumn{1}{r}{\textbf{Coverage}} &
				\multicolumn{1}{|r}{\textbf{MIS}} &
				\multicolumn{1}{r}{\textbf{Coverage}} &
				\multicolumn{1}{|r}{\textbf{MIS}} &
				\multicolumn{1}{r}{\textbf{Coverage}} &
				\multicolumn{1}{|r}{\textbf{MIS}} &
				\multicolumn{1}{r}{\textbf{Coverage}} &
				\multicolumn{1}{|r}{\textbf{MIS}} &
				\multicolumn{1}{r}{\textbf{Coverage}} &
				\multicolumn{1}{|r}{\textbf{MIS}} &
				\multicolumn{1}{r}{\textbf{Coverage}} \\
				\bottomrule\toprule
\textbf{Angola} & \h53.2 & \h94.0 & 3206.0 & 0.0 & 3348.1 & 77.1 & 2089.5 & 0.0 & 29346.8 & 0.0 & 24022.4 & 0.0 & 23844.1 & 0.0 \\
\textbf{Botswana} & \h106.4 & \h93.3 & 4803.4 & 0.0 & 4357.0 & 75.1 & 1706.3 & 10.6 & 53087.1 & 0.0 & 1079.0 & 48.9 & 51432.7 & 0.0 \\
\textbf{Comoros} & \h5.9 & 95.3 & 12.5 & 86.3 & 318.6 & 78.3 & 189.9 & 0.0 & 575.5 & 0.0 & 34.5 & \h100.0 & 225.3 & \h100.0 \\
\textbf{DRC} & \h46.1 & 95.7 & 1725.4 & 0.0 & 1636.2 & 84.0 & 1013.0 & 20.4 & 163.4 & \h100.0 & 336.5 & \h100.0 & 4297.0 & 20.4 \\
\textbf{Eswatini} & \h49.0 & \h90.7 & 2236.0 & 0.0 & 2350.4 & 90.2 & 671.6 & 0.0 & 8795.5 & 42.9 & 7693.3 & 0.0 & 10342.2 & 0.0 \\
\textbf{Lesotho} & \h22.8 & \h97.6 & 795.3 & 0.3 & 1302.8 & 85.2 & 677.5 & 0.0 & 8941.9 & 0.0 & 11845.1 & 0.0 & 1408.6 & 23.3 \\
\textbf{Madagascar} & \h46.5 & 85.4 & 733.9 & 11.8 & 1787.3 & 77.7 & 894.2 & 2.1 & 244.4 & 56.3 & 108.1 & \h100.0 & 1108.4 & \h100.0 \\
\textbf{Malawi} & \h89.4 & 88.8 & 1223.5 & 29.5 & 3526.6 & 85.7 & 1202.7 & 0.0 & 3309.8 & 61.7 & 1153.1 & \h100.0 & 15170.2 & 31.9 \\
\textbf{Mauritius} & \h75.8 & \h74.0 & 468.6 & 0.0 & 484.5 & 79.9 & 454.8 & 0.0 & 6978.1 & 0.0 & 8379.1 & 0.0 & 6861.6 & 0.0 \\
\textbf{Mozambique} & \h66.8 & 94.2 & 244.7 & 91.2 & 1928.7 & 75.0 & 70.9 & 42.1 & 341.8 & \h100.0 & 602.0 & 22.9 & 2707.0 & 22.9 \\
\textbf{Namibia} & 134.0 & 91.3 & 587.2 & 90.8 & 3411.9 & 81.7 & \h101.4 & 89.7 & 21239.2 & 0.0 & 23742.4 & 0.0 & 10490.6 & 0.0 \\
\textbf{Seychelles} & \h5.8 & \h91.0 & 188.7 & 0.0 & 258.6 & 76.1 & 143.5 & 0.0 & 33.4 & 77.6 & 694.2 & 0.0 & 1984.9 & 0.0 \\
\textbf{South Africa} & \h5785.9 & 79.8 & 10713.1 & 97.4 & 97311.6 & 84.7 & 9589.2 & 17.6 & 63626.1 & 48.1 & 33910.2 & \h100.0 & 26489.6 & \h100.0 \\
\textbf{Tanzania} & \h28.4 & \h94.7 & 1853.4 & 0.0 & 1911.1 & 85.9 & 1844.1 & 0.0 & 27048.4 & 0.0 & 35267.8 & 0.0 & 28062.0 & 0.0 \\
\textbf{Zambia} & \h150.0 & 89.7 & 429.3 & 94.9 & 3701.5 & 77.6 & 429.3 & 19.1 & 6940.6 & 6.3 & 458.3 & \h100.0 & 1731.8 & 77.1 \\
\textbf{Zimbabwe} & 169.9 & 87.3 & 810.4 & 85.8 & 5004.7 & 84.6 & \h119.7 & \h100.0 & 26898.9 & 0.0 & 746.0 & 95.8 & 4778.9 & 20.8 \\
				\bottomrule
		\end{tabular}}
	\end{specialtable}
	
	Recall from Section~\ref{subsec:metrics}, the MIS penalizes large intervals, so our model has the narrowest prediction intervals. In Figure~\ref{fig:bar_pi_sadc} we average the results for prediction intervals across the entirety of the SADC region. MES-LSTM consistently has the narrowest intervals for both predictands at all prediction intervals, as evidenced in Figures~\ref{fig:mis_sadc_01},~\ref{fig:mis_sadc_05}, and~\ref{fig:mis_sadc_1}.
	
	In terms of Coverage, our model outperforms the benchmarks for the individual countries except in a few instances. In most of the exceptions, the difference is minuscule, whereas in other instances it is not negligible. What is more important however, is the aggregate performance over the entire region, where MES-LSTM scores better (as can be seen in Figures~\ref{fig:cov_sadc_01},~\ref{fig:cov_sadc_05}, and~\ref{fig:cov_sadc_1}).
	
\end{paracol}
\begin{figure}[htbp]
	\widefigure
	\begin{subfigure}{0.33\columnwidth}
		\includegraphics[width=\columnwidth]{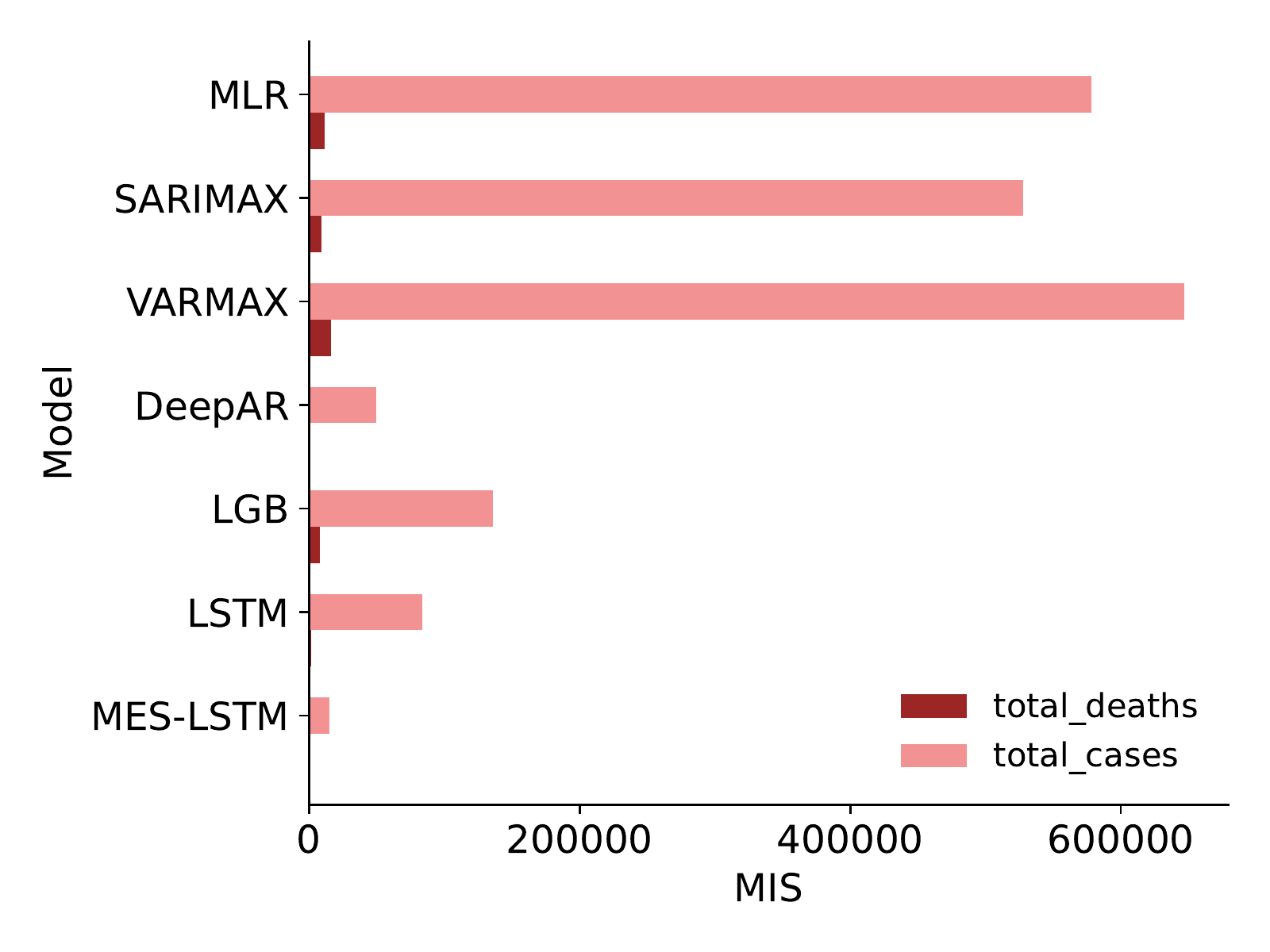}\caption{$\alpha = 0.05$.\label{fig:mis_sadc_01}} 
	\end{subfigure}
	\begin{subfigure}{0.33\columnwidth}
		\includegraphics[width=\columnwidth]{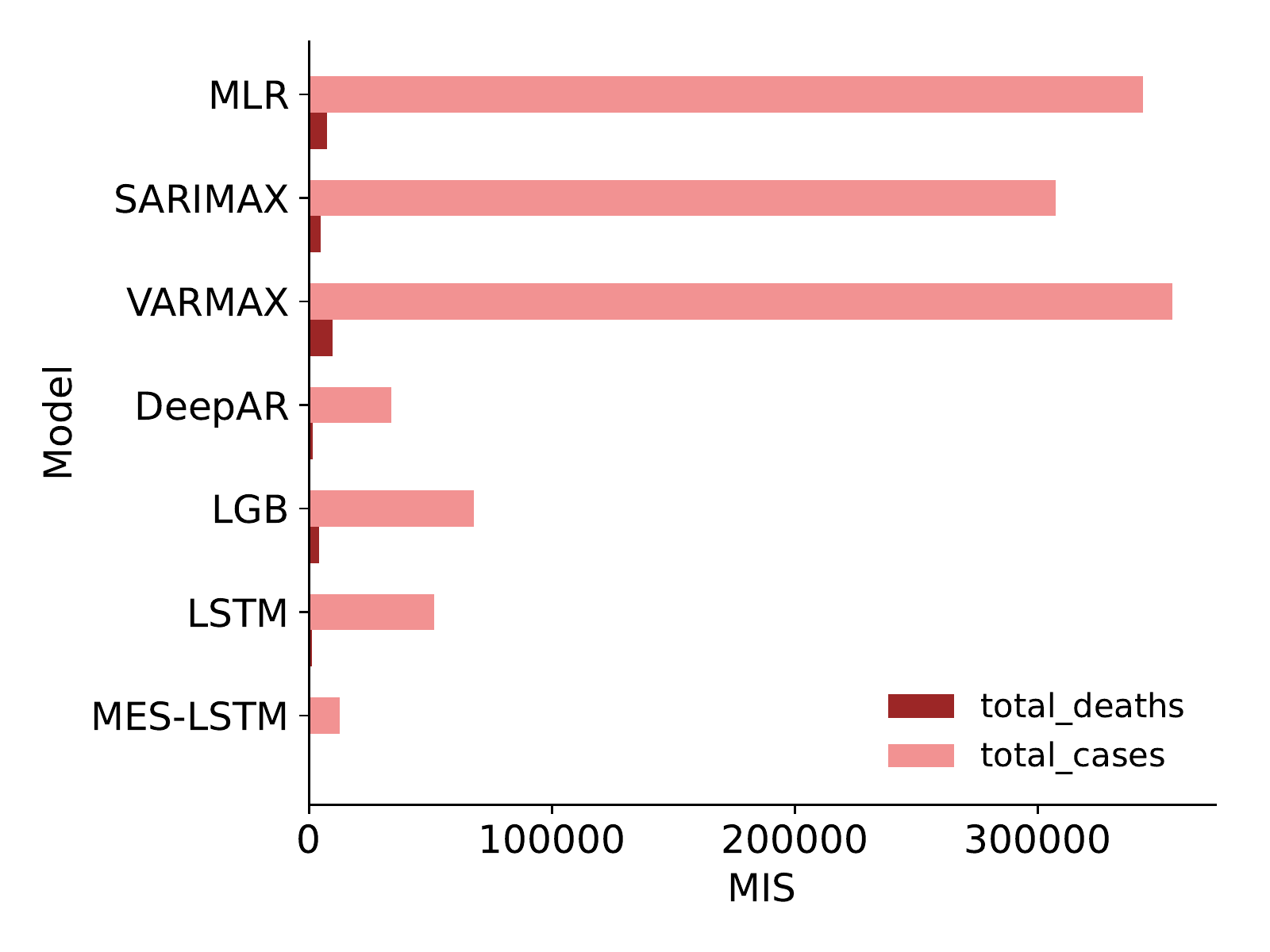}\caption{$\alpha = 0.1$.\label{fig:mis_sadc_05}}
	\end{subfigure}
	\begin{subfigure}{0.33\columnwidth}
		\includegraphics[width=\columnwidth]{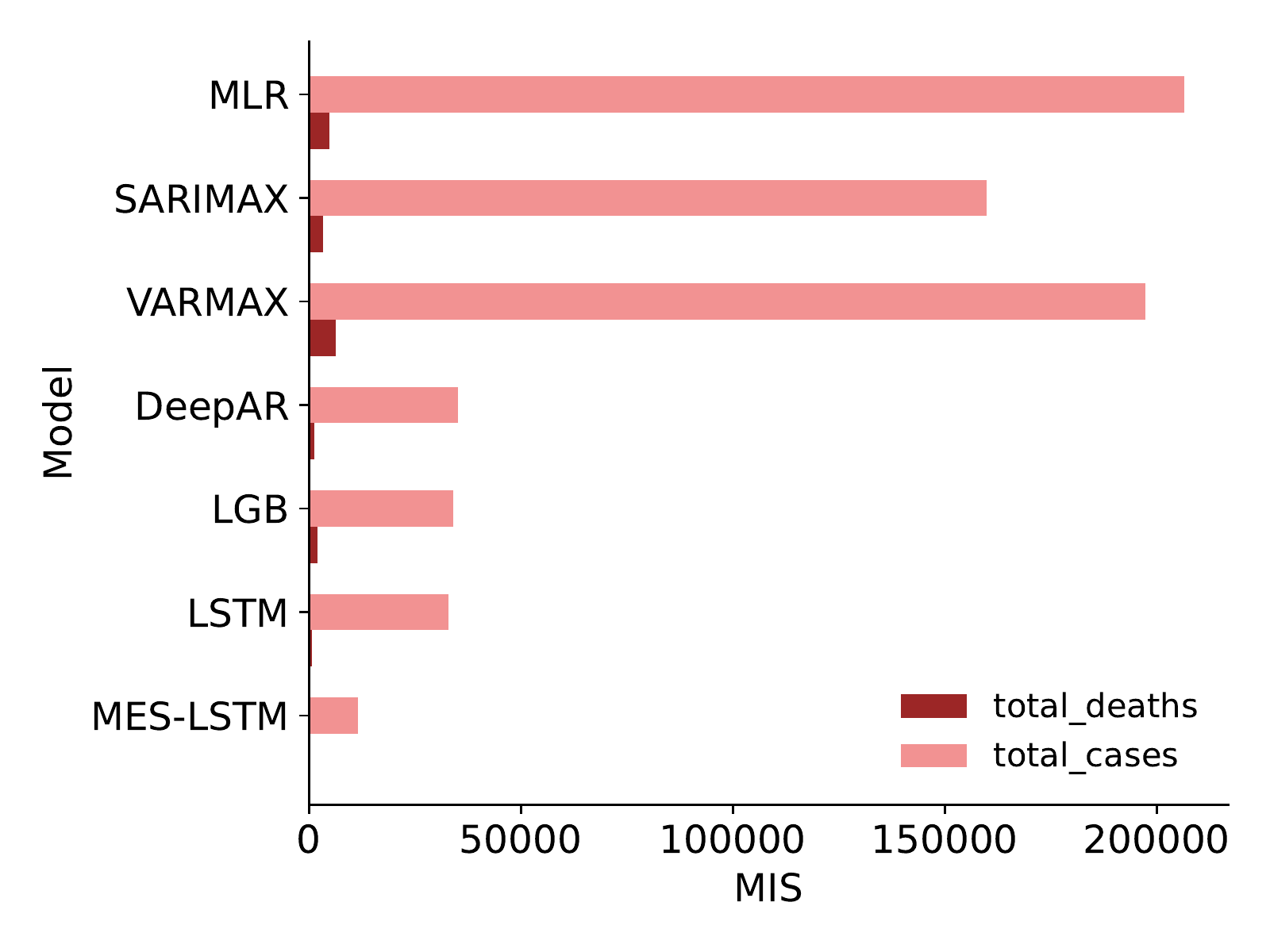}\caption{$\alpha = 0.2$.\label{fig:mis_sadc_1}}
	\end{subfigure}\\ \bigskip
	\begin{subfigure}{0.33\columnwidth}
		\centering 
		\includegraphics[width=\columnwidth]{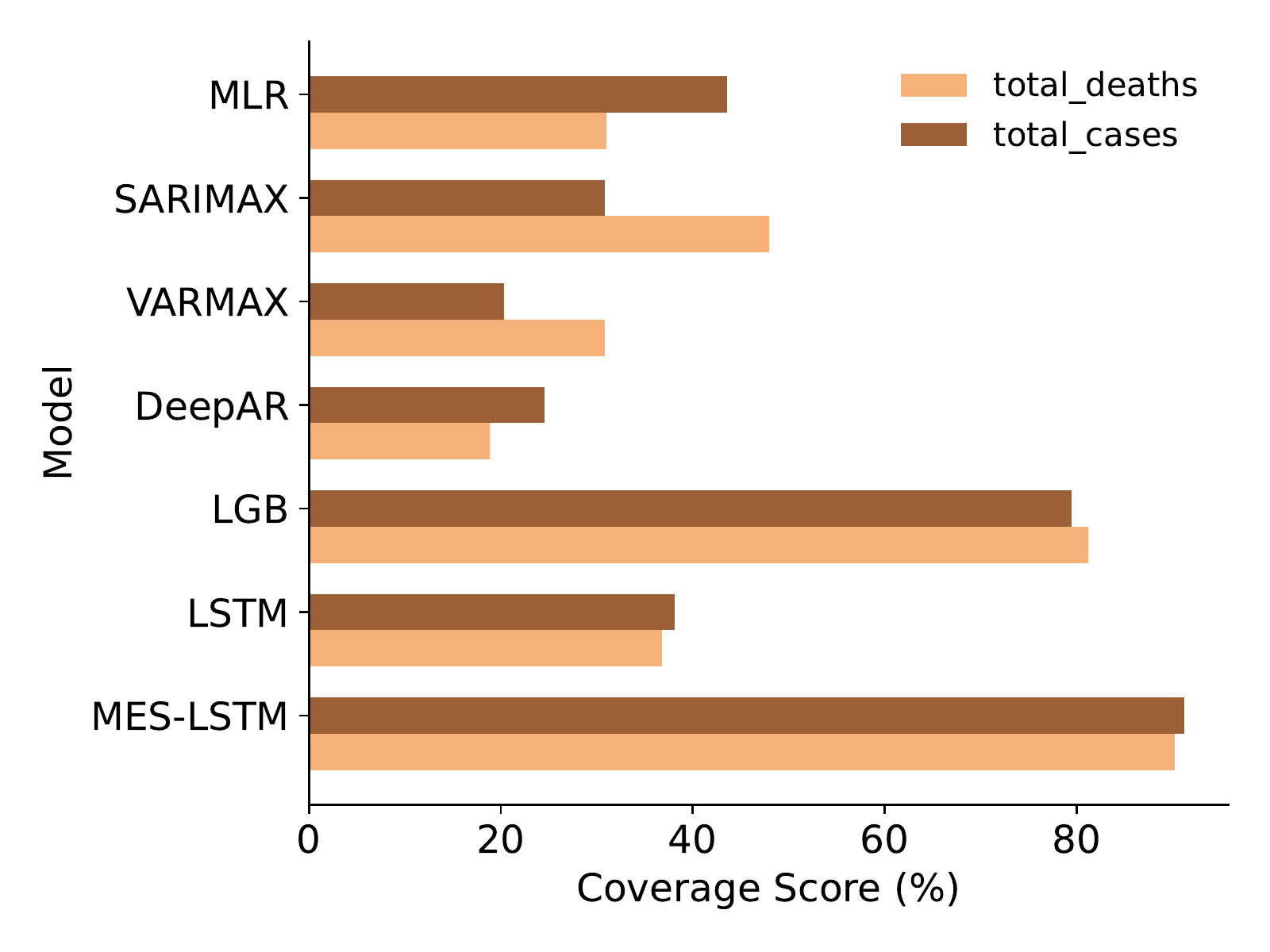}\caption{$\alpha = 0.05$.\label{fig:cov_sadc_01}}
	\end{subfigure}
	\begin{subfigure}{0.33\columnwidth}
		\centering 
		\includegraphics[width=\columnwidth]{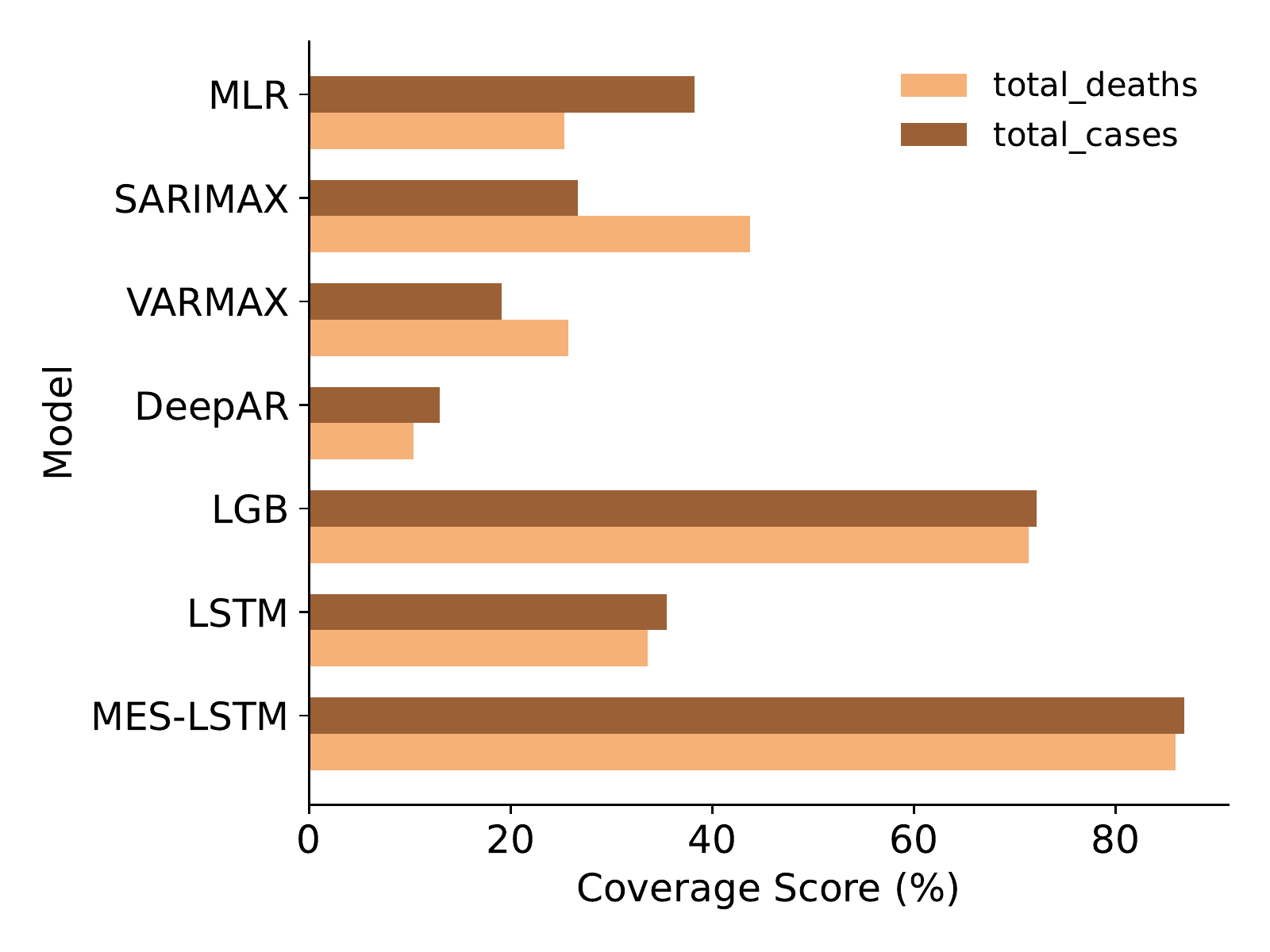}\caption{$\alpha = 0.1$.\label{fig:cov_sadc_05}}
	\end{subfigure}
	\begin{subfigure}{0.33\columnwidth}
		\includegraphics[width=\columnwidth]{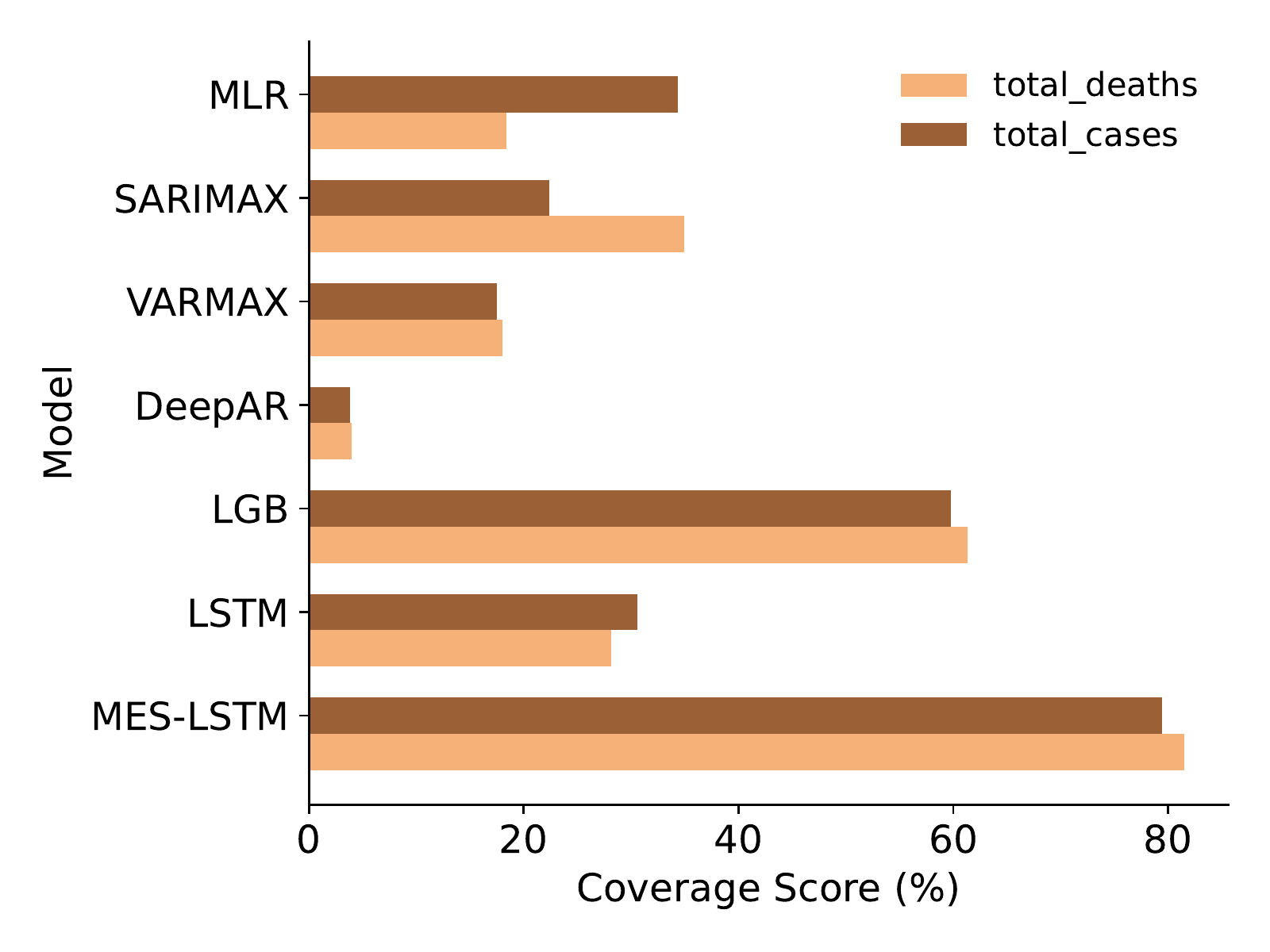}\caption{$\alpha = 0.2$.\label{fig:cov_sadc_1}}
	\end{subfigure}
	\caption{Prediction Interval Accuracy in the SADC Region.\label{fig:bar_pi_sadc}}
\end{figure}
\begin{paracol}{2}
	\switchcolumn
	
	\subsection{Forecast Performance for South Africa}
	After noting the relatively poor performance for South Africa, we believe this requires further probing.
	
	We first present our observed error metrics distribution over the independent trials for all the models for both predictands in South Africa. Tables~\ref{tab:distribution_cases} and~\ref{tab:distribution_deaths} show the forecasting error for \textit{total cases} and \textit{total deaths} respectively. The models VARMAX, SARIMAX and MLR are all deterministic, so the output for different trials is identical. As a result, there is no variation in the accuracy of these models and the standard deviation is zero. For \textit{total cases}, VARMAX (DeepAR) has the lowest (highest) average sMAPE, and MES-LSTM (SARIMAX) has the lowest (highest) average RMSE. For \textit{total deaths}, MES-LSTM has the lowest average sMAPE and RMSE, while and SARIMAX has the highest figures for both.
	
	\begin{specialtable}[h!]
		\centering
		\caption{Forecast Accuracy Distribution for  All Trials for \textit{total cases} in South Africa.}
		\label{tab:distribution_cases}
		\resizebox{1\columnwidth}{!}{%
			\begin{tabular}{lrr|rr|rr|rr|rr|rr|rr}
				\toprule
				&
				\multicolumn{2}{c}{\textbf{MES-LSTM}} &
				\multicolumn{2}{c}{\textbf{LSTM}} &
				\multicolumn{2}{c}{\textbf{LGB}} &

				\multicolumn{2}{c}{\textbf{DeepAR}} &
				\multicolumn{2}{c}{\textbf{VARMAX}} &
				\multicolumn{2}{c}{\textbf{SARIMAX}} &
				\multicolumn{2}{c}{\textbf{MLR}} \\ 
				 &
				\multicolumn{1}{r}{\textbf{sMAPE}} &
				\multicolumn{1}{r}{\textbf{RMSE}} &
				\multicolumn{1}{|r}{\textbf{sMAPE}} &
				\multicolumn{1}{r}{\textbf{RMSE}} &
				\multicolumn{1}{|r}{\textbf{sMAPE}} &
				\multicolumn{1}{r}{\textbf{RMSE}} &
				\multicolumn{1}{|r}{\textbf{sMAPE}} &
				\multicolumn{1}{r}{\textbf{RMSE}} &
				\multicolumn{1}{|r}{\textbf{sMAPE}} &
				\multicolumn{1}{r}{\textbf{RMSE}} &
				\multicolumn{1}{|r}{\textbf{sMAPE}} &
				\multicolumn{1}{r}{\textbf{RMSE}} &
				\multicolumn{1}{|r}{\textbf{sMAPE}} &
				\multicolumn{1}{r}{\textbf{RMSE}} \\
				\bottomrule\toprule
				\textbf{mean} & 5.0 & 26979.2 & 4.8 & 150416.4 & 5.6 & 30420.9 & 7.8 & 212612.0 & 2.8 & 87376.0 & 8.2 & 265998.0 & 4.0 & 118139.5 \\
\textbf{std} & 1.1 & 5641.0 & 3.8 & 121306.0 & 1.3 & 12885.4 & 0.1 & 1192.2 & 0.0 & 0.0 & 0.0 & 0.0 & 0.0 & 0.0 \\
				\bottomrule
		\end{tabular}}
	\end{specialtable}
	
	\begin{specialtable}[h!]
		\centering
		\caption{Forecast Accuracy Distribution for  All Trials for \textit{total deaths} in South Africa.}
		\label{tab:distribution_deaths}
		\resizebox{1\columnwidth}{!}{%
			\begin{tabular}{lrr|rr|rr|rr|rr|rr|rr}
				\toprule
				&
				\multicolumn{2}{c}{\textbf{MES-LSTM}} &
				\multicolumn{2}{c}{\textbf{LSTM}} &
				\multicolumn{2}{c}{\textbf{LGB}} &

				\multicolumn{2}{c}{\textbf{DeepAR}} &
				\multicolumn{2}{c}{\textbf{VARMAX}} &
				\multicolumn{2}{c}{\textbf{SARIMAX}} &
				\multicolumn{2}{c}{\textbf{MLR}} \\ 
				 &
				\multicolumn{1}{r}{\textbf{sMAPE}} &
				\multicolumn{1}{r}{\textbf{RMSE}} &
				\multicolumn{1}{|r}{\textbf{sMAPE}} &
				\multicolumn{1}{r}{\textbf{RMSE}} &
				\multicolumn{1}{|r}{\textbf{sMAPE}} &
				\multicolumn{1}{r}{\textbf{RMSE}} &
				\multicolumn{1}{|r}{\textbf{sMAPE}} &
				\multicolumn{1}{r}{\textbf{RMSE}} &
				\multicolumn{1}{|r}{\textbf{sMAPE}} &
				\multicolumn{1}{r}{\textbf{RMSE}} &
				\multicolumn{1}{|r}{\textbf{sMAPE}} &
				\multicolumn{1}{r}{\textbf{RMSE}} &
				\multicolumn{1}{|r}{\textbf{sMAPE}} &
				\multicolumn{1}{r}{\textbf{RMSE}} \\
				\bottomrule\toprule
				\textbf{mean} & 4.9 & 446.2 & 7.8 & 7902.4 & 5.9 & 2773.4 & 10.4 & 8515.8 & 10.7 & 10806.6 & 16.1 & 15438.9 & 6.5 & 6276.2 \\
\textbf{std} & 1.2 & 106.0 & 5.3 & 5578.8 & 2.2 & 1259.7 & 0.1 & 61.2 & 0.0 & 0.0 & 0.0 & 0.0 & 0.0 & 0.0 \\
				\bottomrule
		\end{tabular}}
	\end{specialtable}
	
	MES-LSTM and SARIMAX results being on opposite sides of the spectrum is also evidenced by the box and whisker plots in Figure~\ref{fig:box}.
	
	\begin{figure}[h!tbp]
		\begin{subfigure}{0.45\columnwidth}
			\includegraphics[width=\columnwidth]{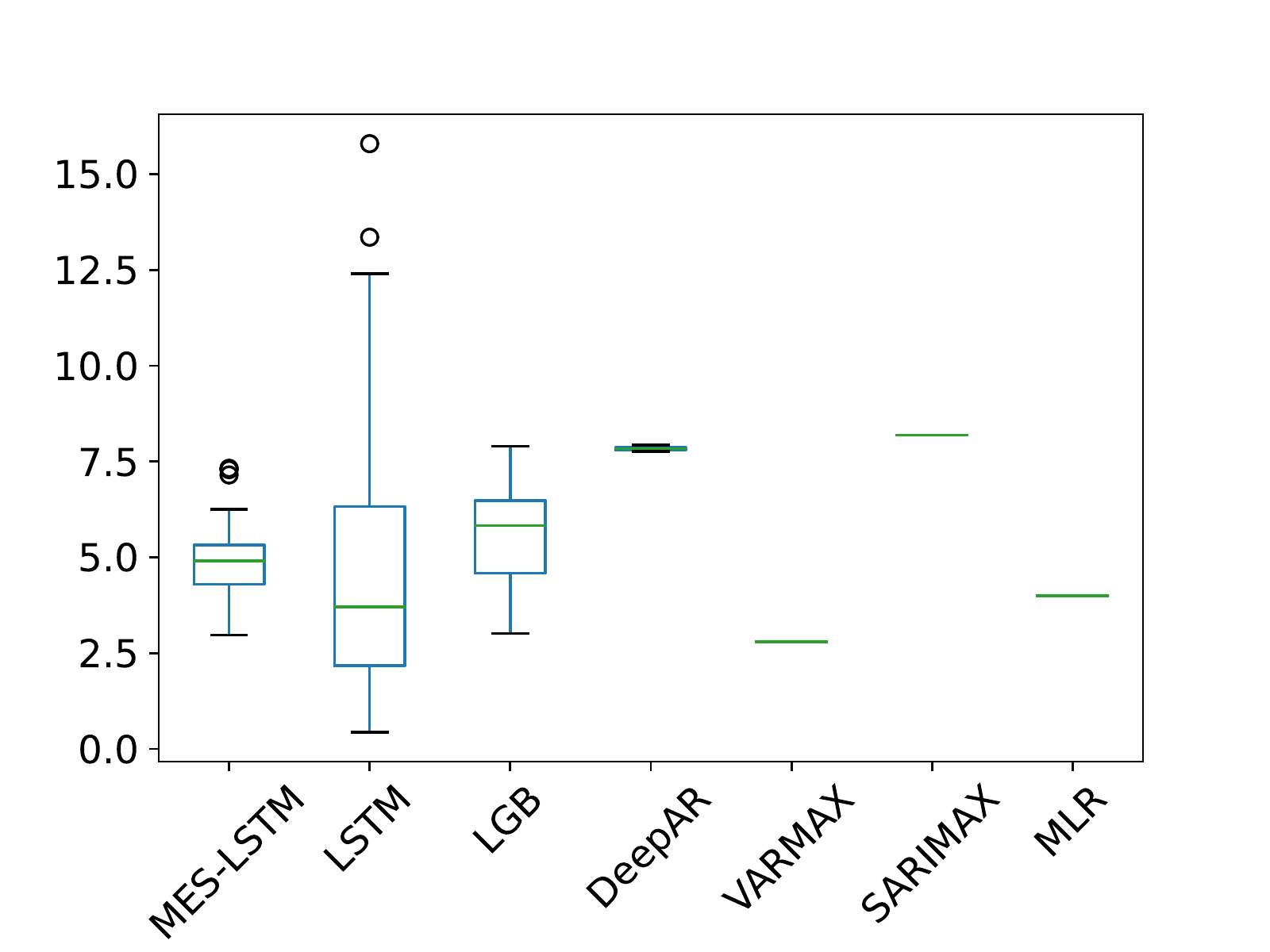}\caption{sMAPE for \textit{total cases}.\label{fig:smape_box_cases}} 
		\end{subfigure}
		\begin{subfigure}{0.45\columnwidth}
			\includegraphics[width=\columnwidth]{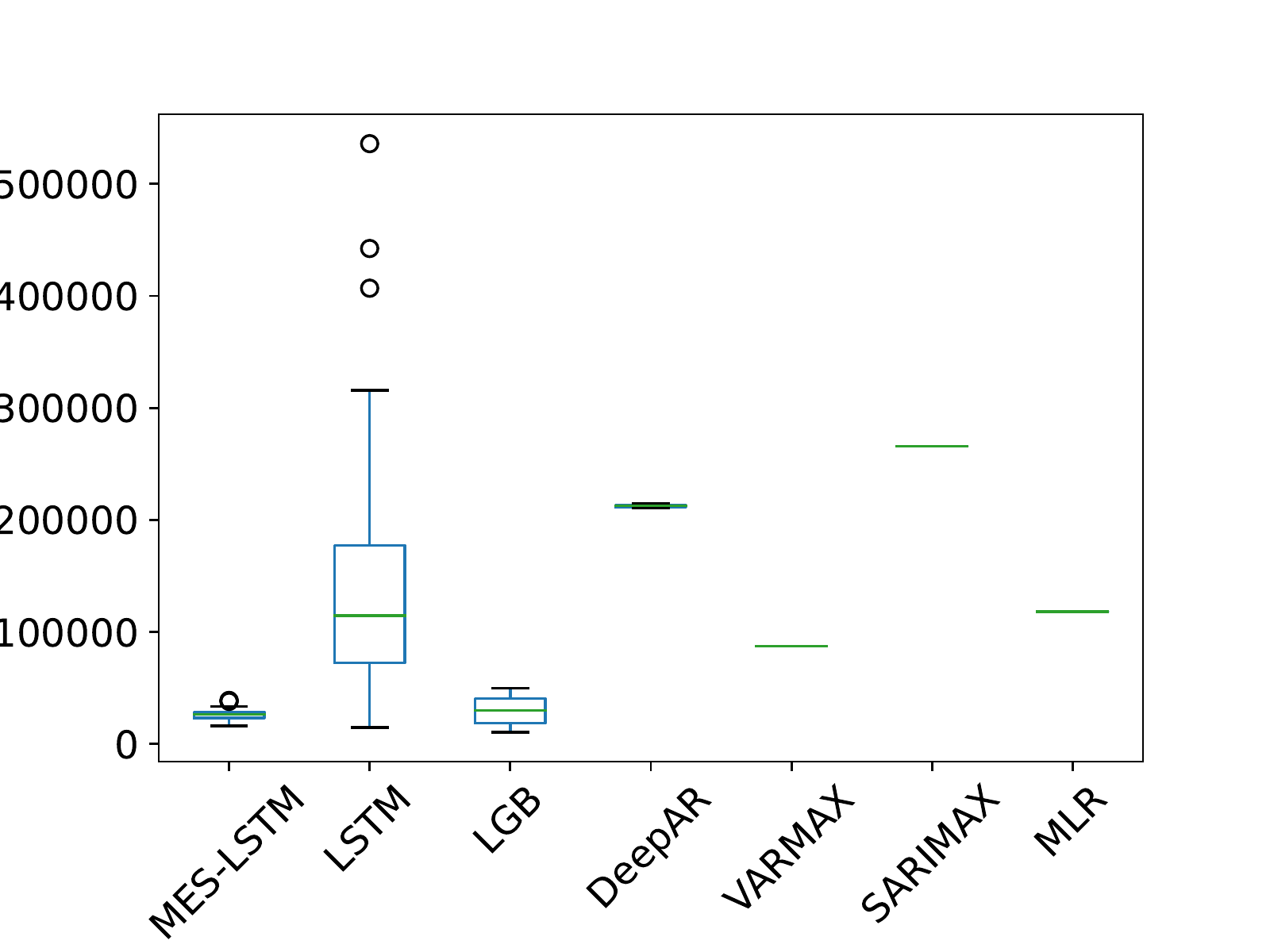}\caption{RMSE for \textit{total cases}.\label{fig:rmse_box_cases}}
		\end{subfigure}\\ \vfill
		\begin{subfigure}{0.45\columnwidth}
			\centering 
			\includegraphics[width=\columnwidth]{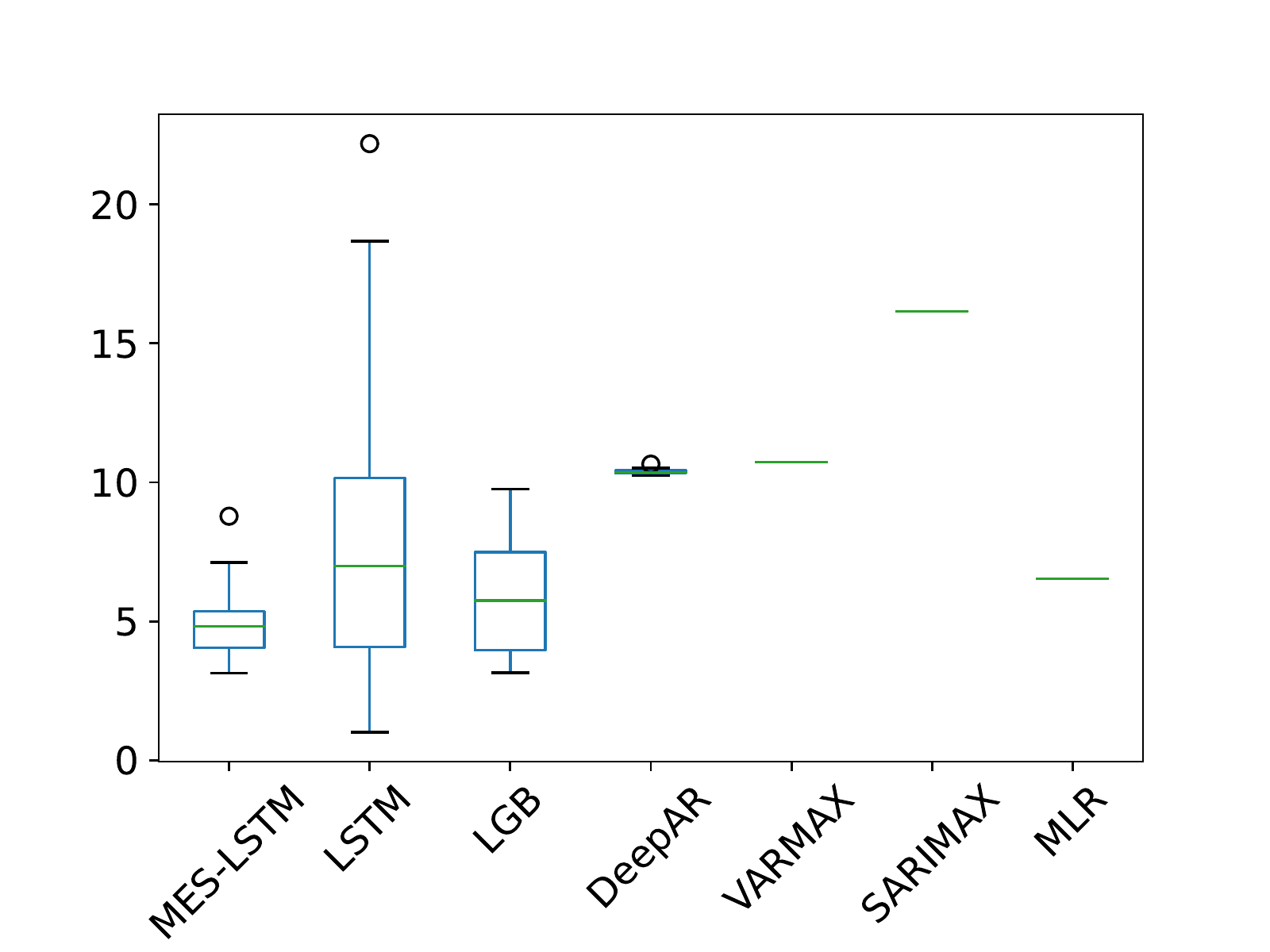}\caption{sMAPE for \textit{total deaths}.\label{fig:smape_box_deaths}}
		\end{subfigure}
		\begin{subfigure}{0.45\columnwidth}
			\centering 
			\includegraphics[width=\columnwidth]{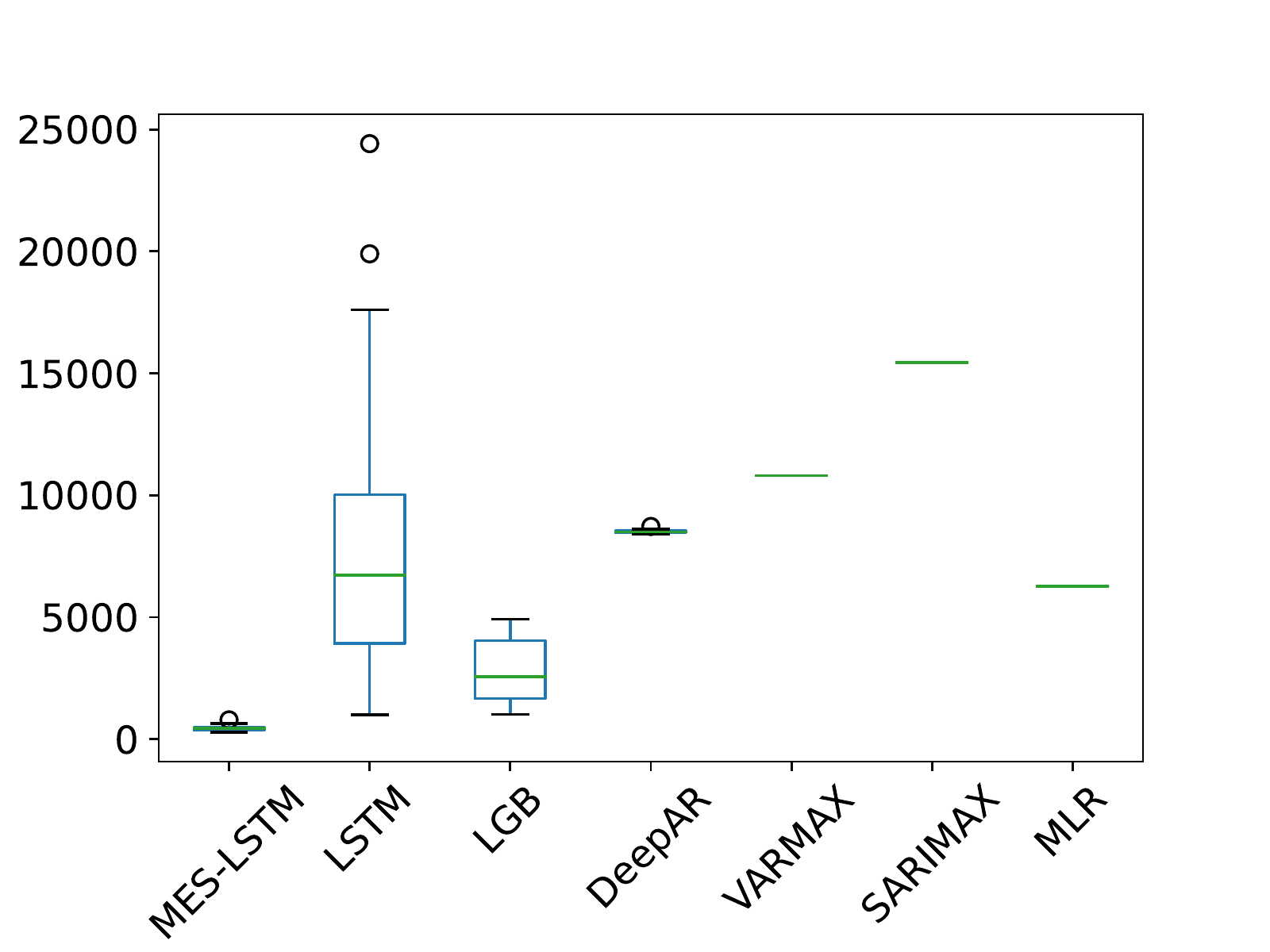}\caption{RMSE for \textit{total deaths}.\label{fig:rmse_box_deaths}}
		\end{subfigure}
		\caption{Forecast Accuracy Distribution Boxplots for  All Trials in South Africa.\label{fig:box}}
	\end{figure}
	
	The results from the LSTM are far more variable for each trial than MES-LSTM. LSTM seems competitive to our model for some trials but shows little consistency for multiple independent runs. The LSTM's relatively poor and  oft-inconsistent performance with COVID-19 modeling agrees with observations by other researchers (as detailed in Section~\ref{sub:litreview}.)
	
	When viewed together with the bar graphs presented in Figure~\ref{fig:bar_forecast_sa}, one possible intepretation of all the forecast results is a ranking of the best performance for South Africa in increasing order as SARIMAX, VARMAX, LSTM, DeepAR, MLR, LGB, and MES-LSTM. MES-LSTM shows consistent outperformance over (or at least competitiveness to) the benchmarks. Our model also presents results with a tight distribution, second in terms of tightness to DeepAR (from the probabilistic forecasts models). Even in instances where outliers are present in the forecast distribution of our model, these outliers are still very close to the core distribution. This tendency reaffirms the robustness of our model when it comes to producing accurate forecasts.
	
	\begin{figure}[h!tbp]
		\centering
		\resizebox{\columnwidth}{!}{
			\begin{subfigure}{0.49\columnwidth}
				\includegraphics[width=\columnwidth]{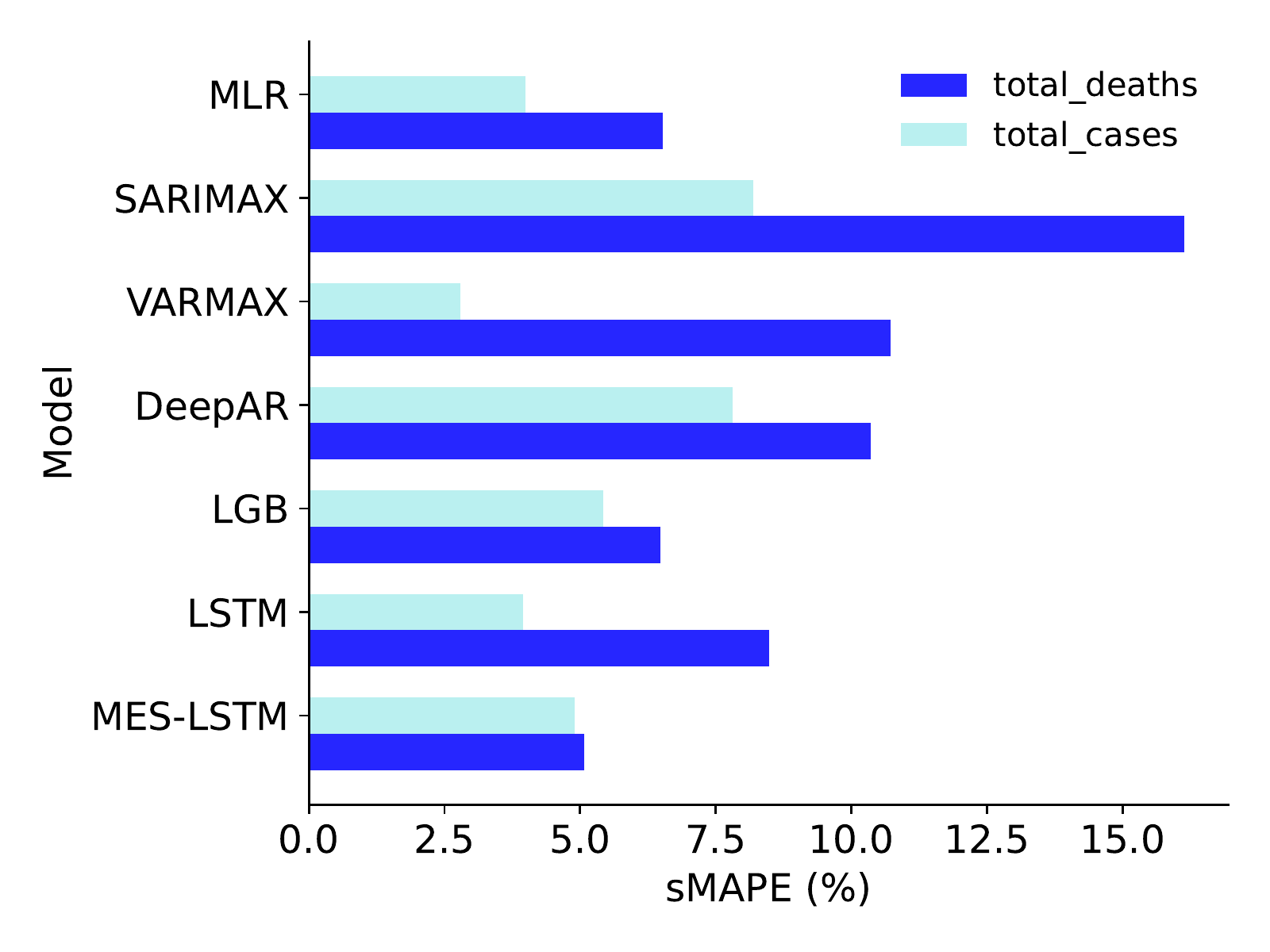}
				\caption{sMAPE}
				\label{fig:smape_bar_sa}
			\end{subfigure}
			\begin{subfigure}{0.49\columnwidth}
				\includegraphics[width=\columnwidth]{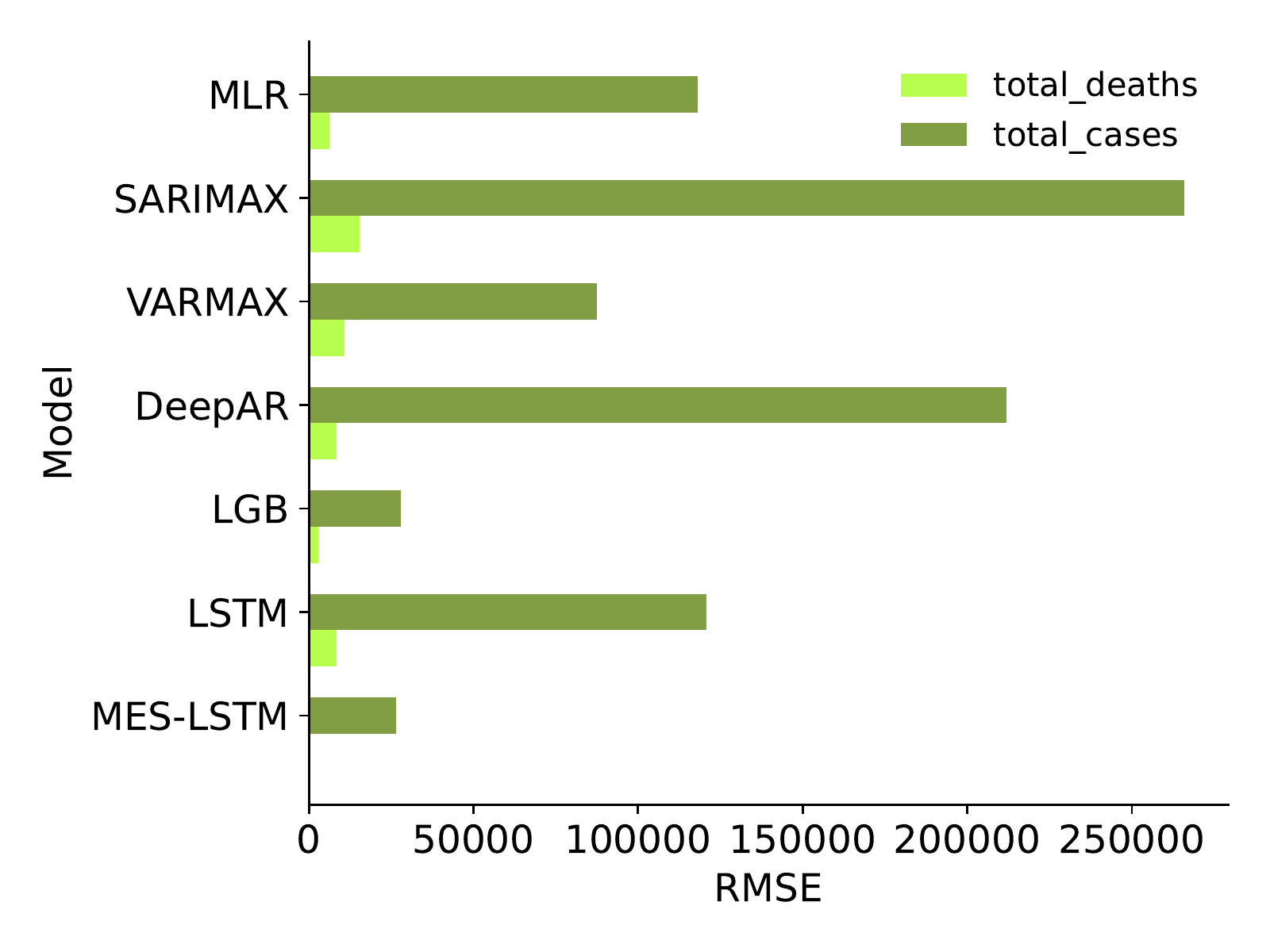}
				\caption{RMSE.}
				\label{fig:rmse_bar_sa}
		\end{subfigure}}
		\caption{Forecast Accuracy in South Africa.\label{fig:bar_forecast_sa}}
	\end{figure}
	
	\subsection{Prediction Interval Performance for South Africa}
	We turn our attention to the prediction interval accuracy for South Africa. Tables~\ref{tab:distribution_cases_01} and~\ref{tab:distribution_deaths_01} show the prediction interval summary statistics at the $\alpha = 0.05$ level of significance for our respective predictands. We note that the distribution for MES-LSTM is tight and this characteristic persists throughout all significance levels.
	
	\begin{specialtable}[h!]
		\centering
		\caption{Prediction Interval Accuracy Distribution for All Trials for \textit{total cases} in South Africa ($\alpha = 0.05$).}
		\label{tab:distribution_cases_01}
		\resizebox{1\columnwidth}{!}{%
			\begin{tabular}{lrr|rr|rr|rr|rr|rr|rr}
				\toprule
				& \multicolumn{2}{c}{\textbf{MES-LSTM}} & \multicolumn{2}{c}{\textbf{LSTM}} & \multicolumn{2}{c}{\textbf{LGB}} & \multicolumn{2}{c}{\textbf{DeepAR}} &\multicolumn{2}{c}{\textbf{VARMAX}} & \multicolumn{2}{c}{\textbf{SARIMAX}} & \multicolumn{2}{c}{\textbf{MLR}} \\ 
				& \multicolumn{1}{c}{\textbf{MIS}} & \textbf{Coverage} & \multicolumn{1}{c}{\textbf{MIS}} & \textbf{Coverage} & \multicolumn{1}{c}{\textbf{MIS}} & \textbf{Coverage} &	\multicolumn{1}{c}{\textbf{MIS}} & \textbf{Coverage} & \multicolumn{1}{c}{\textbf{MIS}} & \textbf{Coverage} & \multicolumn{1}{c}{\textbf{MIS}} & \textbf{Coverage} & \multicolumn{1}{c}{\textbf{MIS}} & \textbf{Coverage} \\
				\bottomrule\toprule
				\textbf{mean} & 193075.6 & 80.7 & 348860.7 & 97.7 & 1418713.2 & 86.6 & 239009.3 & 100.0 & 740131.9 & 100.0 & 1061162.0 & 100.0 & 828447.2 & 100.0 \\
\textbf{std} & 13881.3 & 11.2 & 76242.6 & 13.7 & 0.0 & 30.2 & 1674.9 & 0.0 & 0.0 & 0.0 & 0.0 & 0.0 & 0.0 & 0.0 \\
				\bottomrule
		\end{tabular}}
	\end{specialtable}
	
	\begin{specialtable}[h!]
		\centering
		\caption{Prediction Interval Accuracy Distribution for  All Trials for \textit{total deaths} in South Africa ($\alpha = 0.05$).}
		\label{tab:distribution_deaths_01}
		\resizebox{1\columnwidth}{!}{%
			\begin{tabular}{lrr|rr|rr|rr|rr|rr|rr}
				\toprule
				& \multicolumn{2}{c}{\textbf{MES-LSTM}} & \multicolumn{2}{c}{\textbf{LSTM}} & \multicolumn{2}{c}{\textbf{LGB}} & \multicolumn{2}{c}{\textbf{DeepAR}} &\multicolumn{2}{c}{\textbf{VARMAX}} & \multicolumn{2}{c}{\textbf{SARIMAX}} & \multicolumn{2}{c}{\textbf{MLR}} \\ 
				& \multicolumn{1}{c}{\textbf{MIS}} & \textbf{Coverage} & \multicolumn{1}{c}{\textbf{MIS}} & \textbf{Coverage} & \multicolumn{1}{c}{\textbf{MIS}} & \textbf{Coverage} &	\multicolumn{1}{c}{\textbf{MIS}} & \textbf{Coverage} & \multicolumn{1}{c}{\textbf{MIS}} & \textbf{Coverage} & \multicolumn{1}{c}{\textbf{MIS}} & \textbf{Coverage} & \multicolumn{1}{c}{\textbf{MIS}} & \textbf{Coverage} \\
				\bottomrule\toprule
				\textbf{mean} & 5785.9 & 79.8 & 10713.1 & 97.4 & 97311.6 & 84.7 & 9589.2 & 17.6 & 63626.1 & 48.1 & 33910.2 & 100.0 & 26489.6 & 100.0 \\
\textbf{std} & 1714.1 & 15.1 & 3448.0 & 15.6 & 0.0 & 28.5 & 723.9 & 6.5 & 0.0 & 0.0 & 0.0 & 0.0 & 0.0 & 0.0 \\
				\bottomrule
			\end{tabular}
		}
	\end{specialtable}
	
	Examining the bar graphs in Figure~\ref{fig:bar_pi_sa}, we note that even though our model's coverage score is outperformed in some instances, it is not by a great margin. Furthermore, our model has the benefit of the lowest MIS, indicating the tightest prediction intervals. The second benefit is consistency, i.e. where other models may only have decent coverage for one predictantd, our model has a consistent level of coverage across all the predictands. Moreover, we can depend on the significance level, with stricter levels leading to marginally higher coverage. This coverage-alpha dependence from MES-LSTM is important as it also speaks to the robustness of our model. In contrast, DeepAR for instance, goes from almost complete coverage ($\alpha = 0.05, \, \alpha = 0.1$) for \textit{total cases} to none ($\alpha = 0.02$) and MLR presents close to perfect coverage for \textit{total cases} throughout. This does not instill much trust in the statistical method's uncertainty quantification.
	
\end{paracol}
\begin{figure}[htbp]
	\widefigure
	\begin{subfigure}{0.33\columnwidth}
		\includegraphics[width=\columnwidth]{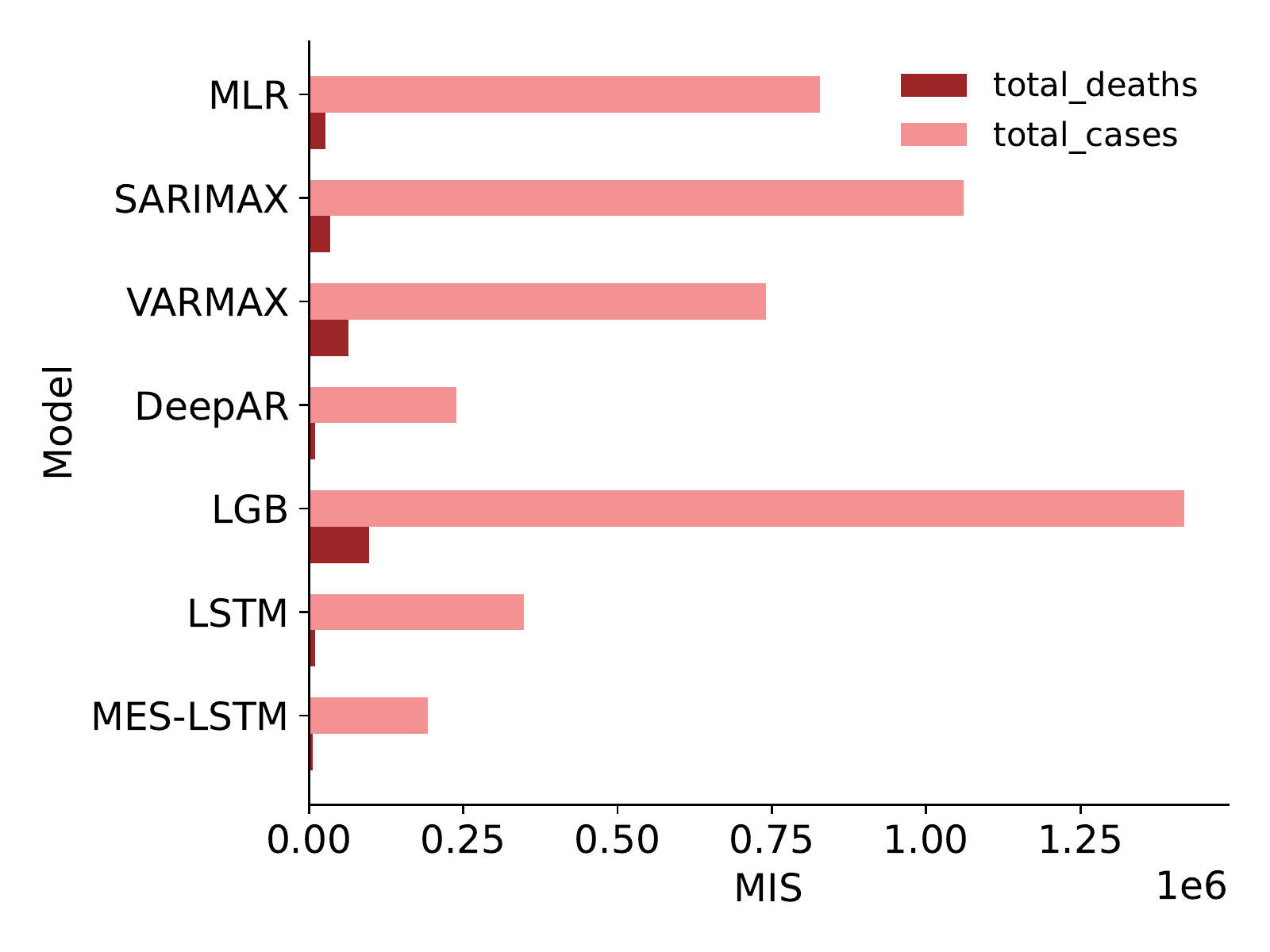}\caption{$\alpha = 0.05$.\label{fig:mis_sa_01}} 
	\end{subfigure}
	\begin{subfigure}{0.33\columnwidth}
		\includegraphics[width=\columnwidth]{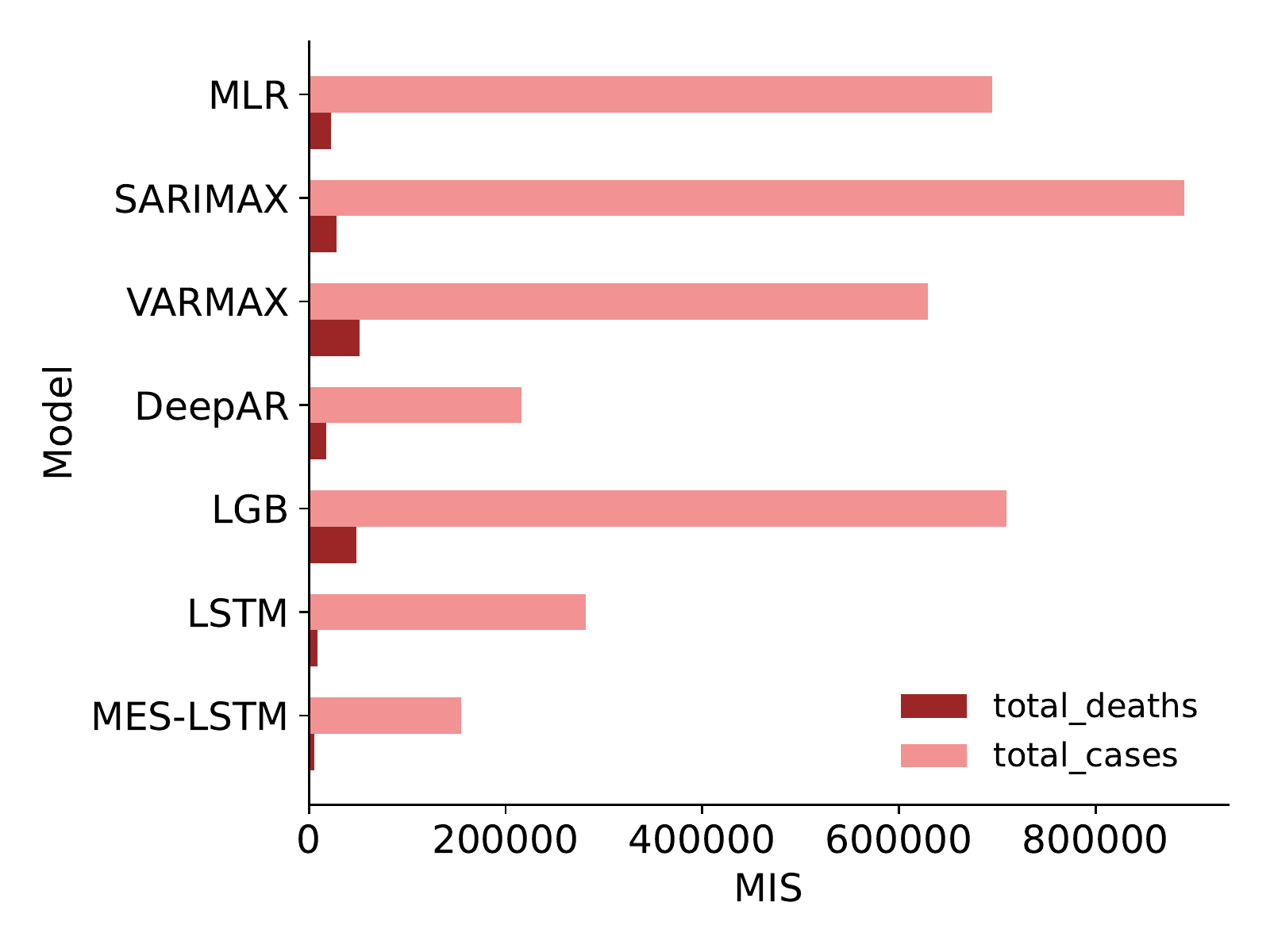}\caption{$\alpha = 0.1$.\label{fig:mis_sa_05}}
	\end{subfigure}
	\begin{subfigure}{0.33\columnwidth}
		\includegraphics[width=\columnwidth]{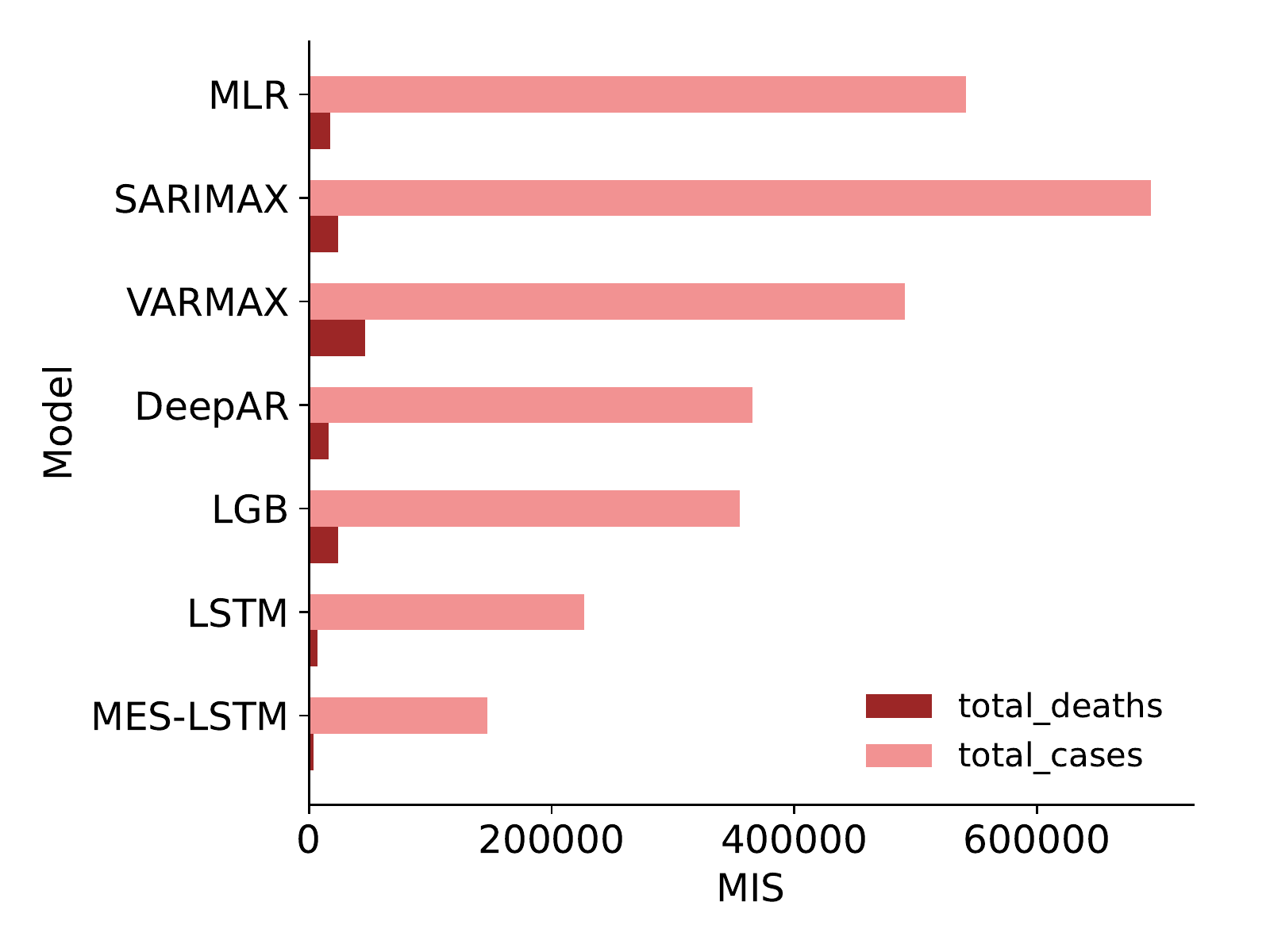}\caption{$\alpha = 0.2$.\label{fig:mis_sa_1}}
	\end{subfigure}\\ \bigskip
	\begin{subfigure}{0.33\columnwidth}
		\centering 
		\includegraphics[width=\columnwidth]{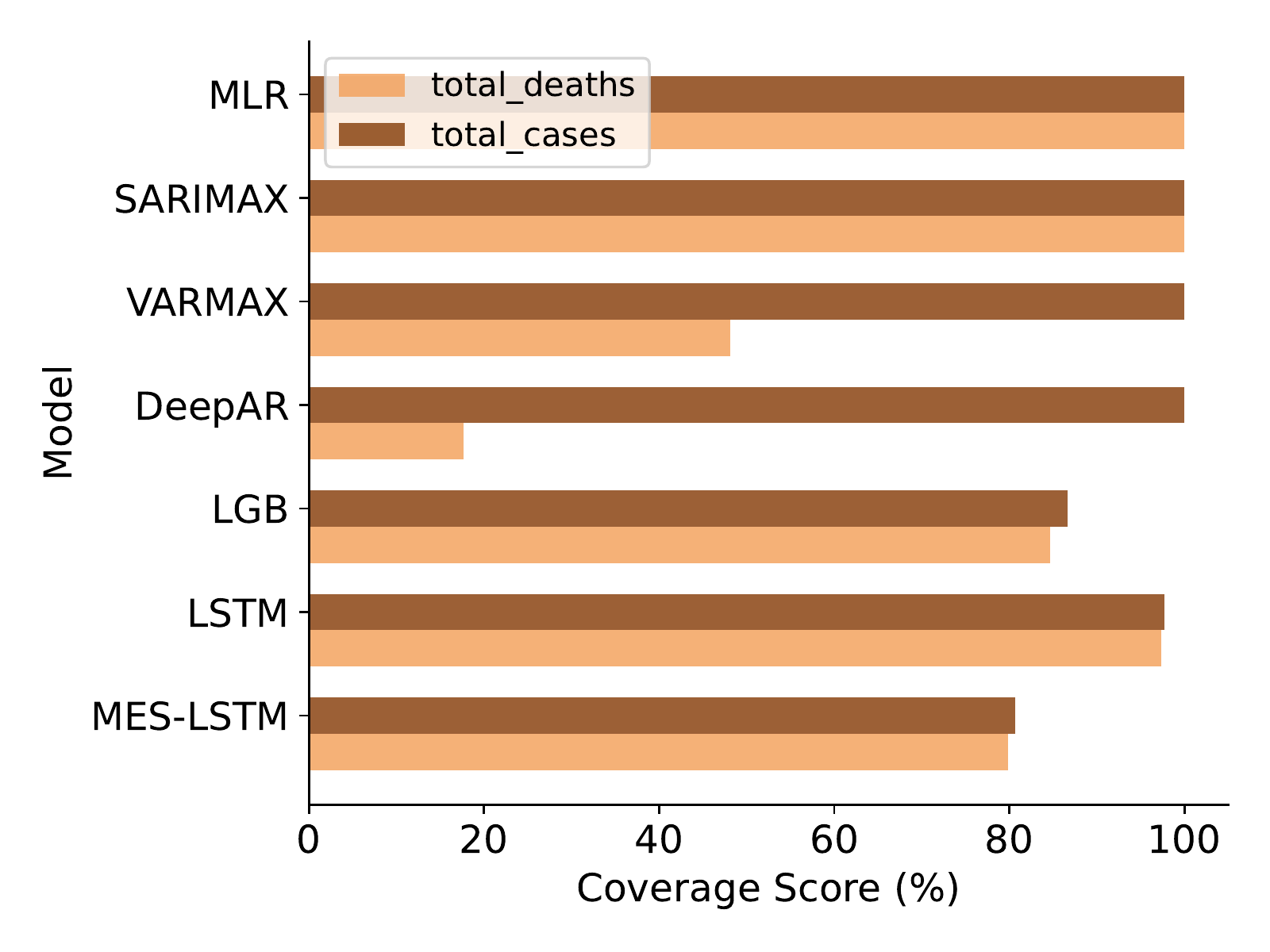}\caption{$\alpha = 0.05$.\label{fig:cov_sa_01}}
	\end{subfigure}
	\begin{subfigure}{0.33\columnwidth}
		\centering 
		\includegraphics[width=\columnwidth]{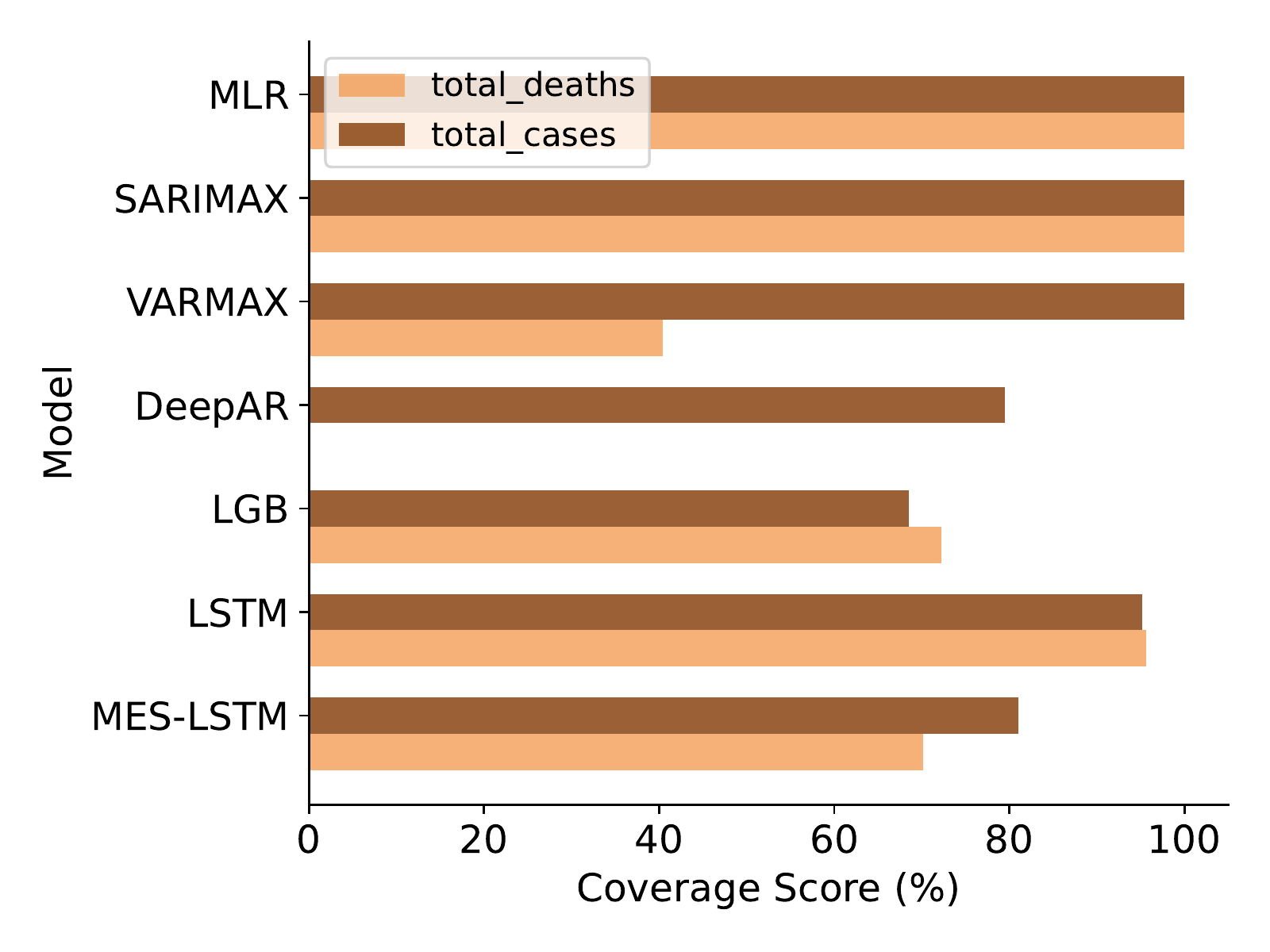}\caption{$\alpha = 0.1$.\label{fig:cov_sa_05}}
	\end{subfigure}
	\begin{subfigure}{0.33\columnwidth}
		\includegraphics[width=\columnwidth]{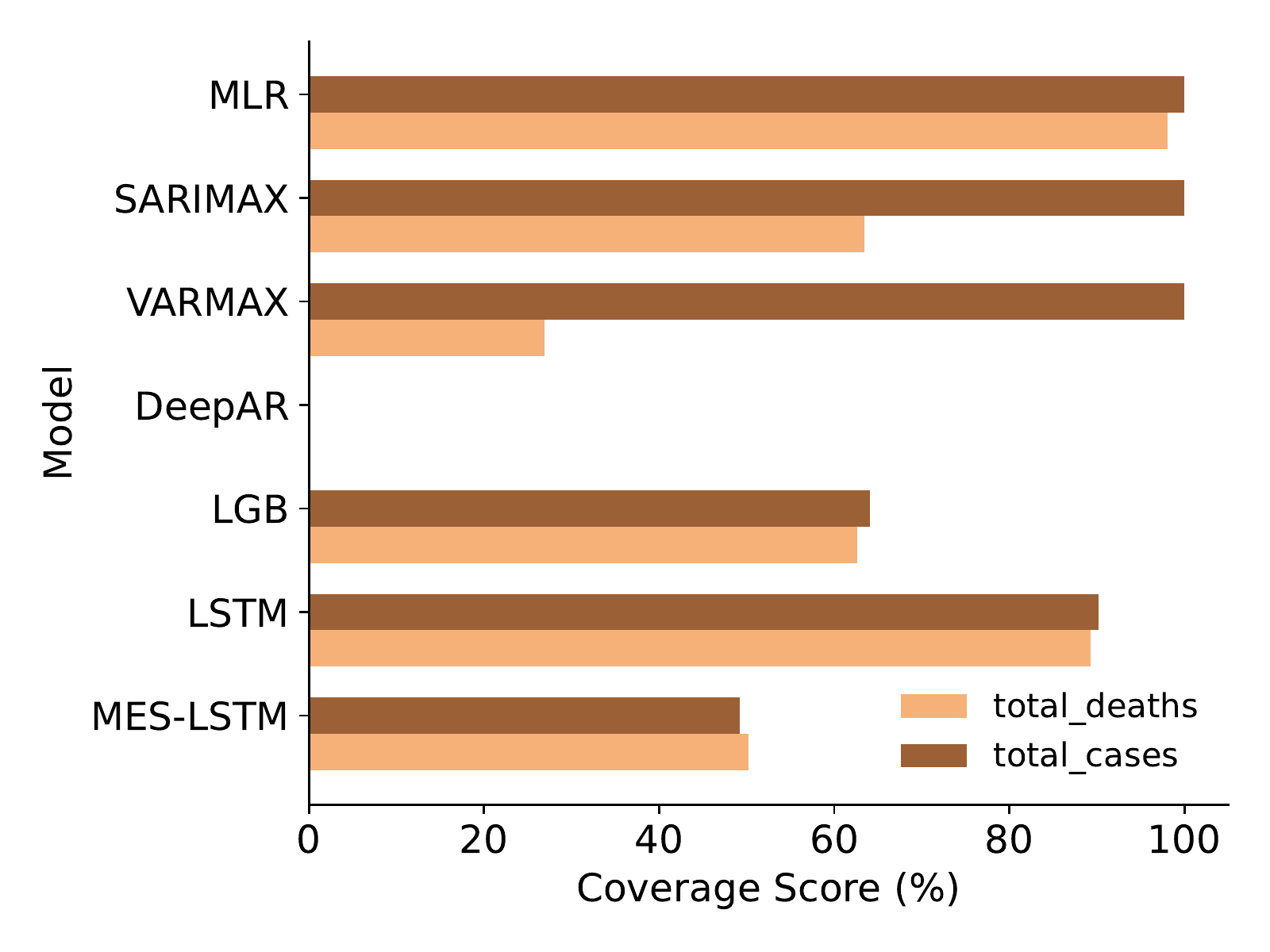}\caption{$\alpha = 0.2$.\label{fig:cov_sa_1}}
	\end{subfigure}
	\caption{Prediction Interval Accuracy in South Africa.\label{fig:bar_pi_sa}}
\end{figure}
\begin{paracol}{2}
	\switchcolumn
	
	Overall, all models perform worse for the SADC region than for South Africa. The worse performance is due to more variability introduced in the data and thus a higher level of uncertainty. This variability can be seen when comparing the Coverage scores from Figures~\ref{fig:bar_pi_sa} to those we presented in Figure~\ref{fig:bar_pi_sadc}.
	
	As an additional point, we perform a one-sided t-test to check whether or not the distribution of our model's forecast and prediction interval results are significantly better than those produced by the benchmark models. Concisely, we compare MES-LSTM, to each of the other models in turn. The t-test results are presented in Tables~\ref{tab:sig_rmse} -~\ref{tab:sig_smape} and Tables~\ref{tab:sig_mis} -~\ref{tab:sig_cov} truncated to three decimal places. Our null hypothesis is H$_0$: The benchmark models produce forecasts that are more accurate and prediction intervals superior to those produced by MES-LSTM.
	
	\begin{specialtable}[htbp]
		\centering
		\caption{Student's t-test for Forecast RMSE (H$_0$: Benchmark Forecasts Outperform MES-LSTM.)}
		\label{tab:sig_rmse}
		\resizebox{1\columnwidth}{!}{%
			\begin{tabular}{@{}lrrrrrr|rrrrrr@{}}
				\toprule
				& \multicolumn{6}{c}{{\textbf{Total Cases}}} & \multicolumn{6}{c}{{\textbf{Total Deaths}}} \\ 
				& \textbf{LSTM} &
				\textbf{LGB} &
				\textbf{DeepAR} &
				\textbf{VARMAX} & \textbf{SARIMAX} & \textbf{MLR} & \textbf{LSTM} &
				\textbf{LGB} &
				\textbf{DeepAR} &
				\textbf{VARMAX} & \textbf{SARIMAX} & \textbf{MLR} \\ \bottomrule\toprule
				\textbf{statistic} & -6.014 & \h-1.448 & -190.477 & -63.342 & -250.673 & -95.605 & -7.906 & -10.891 & -389.954 & -578.058 & -836.515 & -325.283 \\
\textbf{p-value} & 0.000 & \h0.077 & 0.000 & 0.000 & 0.000 & 0.000 & 0.000 & 0.000 & 0.000 & 0.000 & 0.000 & 0.000 \\ \bottomrule
		\end{tabular}}
	\end{specialtable}
	
	\begin{specialtable}[htbp]
		\centering
		\caption{Student's t-test for Forecast sMAPE (H$_0$: Benchmark Forecasts Outperform MES-LSTM.)}
		\label{tab:sig_smape}
		\resizebox{1\columnwidth}{!}{%
			\begin{tabular}{@{}lrrrrrr|rrrrrr@{}}
				\toprule
				& \multicolumn{6}{c}{{\textbf{Total Cases}}} & \multicolumn{6}{c}{{\textbf{Total Deaths}}} \\ 
				& \textbf{LSTM} &
				\textbf{LGB} &
				\textbf{DeepAR} &
				\textbf{VARMAX} & \textbf{SARIMAX} & \textbf{MLR} & \textbf{LSTM} &
				\textbf{LGB} &
				\textbf{DeepAR} &
				\textbf{VARMAX} & \textbf{SARIMAX} & \textbf{MLR} \\ \bottomrule\toprule
				\textbf{statistic} & \h0.270 & \h-1.883 & -15.455 & \h12.180 & -17.383 & \h5.616 & -3.216 & -2.422 & -27.328 & -29.110 & -56.140 & -8.166 \\
\textbf{p-value} & \h0.606 & \h0.032 & 0.000 & \h1.000 & 0.000 & \h1.000 & 0.001 & 0.009 & 0.000 & 0.000 & 0.000 & 0.000 \\ \bottomrule
		\end{tabular}}
	\end{specialtable}

	\begin{specialtable}[htbp]
		\centering
		\caption{Student's t-test for Prediction Interval MIS (H$_0$: Benchmark Prediction Intervals Outperform MES-LSTM.)}
		\label{tab:sig_mis}
		\resizebox{1\columnwidth}{!}{%
			\begin{tabular}{@{}lrrrrrr|rrrrrr@{}}
				\toprule
				& \multicolumn{6}{c}{{\textbf{Total Cases}}} & \multicolumn{6}{c}{{\textbf{Total Deaths}}} \\ 
				& \textbf{LSTM} &
				\textbf{LGB} &
				\textbf{DeepAR} &
				\textbf{VARMAX} & \textbf{SARIMAX} & \textbf{MLR} & \textbf{LSTM} &
				\textbf{LGB} &
				\textbf{DeepAR} &
				\textbf{VARMAX} & \textbf{SARIMAX} & \textbf{MLR} \\ \bottomrule\toprule
				\textbf{statistic} & -5.328 & -25.791 & -2.851 & -22.095 & -34.220 & -25.119 & -2.299 & -37.924 & -10.070 & -40.222 & -20.029 & -14.501 \\
\textbf{p-value} & 0.000 & 0.000 & 0.004 & 0.000 & 0.000 & 0.000 & 0.013 & 0.000 & 0.000 & 0.000 & 0.000 & 0.000 \\ \bottomrule
		\end{tabular}}
	\end{specialtable}
	
	\begin{specialtable}[htbp]
		\centering
		\caption{Student's t-test for Prediction Interval Coverage (H$_0$: Benchmark Prediction Intervals Outperform MES-LSTM.)}
		\label{tab:sig_cov}
		\resizebox{1\columnwidth}{!}{%
			\begin{tabular}{@{}lrrrrrr|rrrrrr@{}}
				\toprule
				& \multicolumn{6}{c}{{\textbf{Total Cases}}} & \multicolumn{6}{c}{{\textbf{Total Deaths}}} \\ 
				& \textbf{LSTM} &
				\textbf{LGB} &
				\textbf{DeepAR} &
				\textbf{VARMAX} & \textbf{SARIMAX} & \textbf{MLR} & \textbf{LSTM} &
				\textbf{LGB} &
				\textbf{DeepAR} &
				\textbf{VARMAX} & \textbf{SARIMAX} & \textbf{MLR} \\ \bottomrule\toprule
				\textbf{statistic} & \h-1.8770 & \h1.6106 & \h0.2267 & \h-2.7459 & \h-2.7459 & \h-2.7459 & \h-2.8639 & \h-0.2274 & 8.3710 & 3.5521 & \h-3.5613 & \h-3.5613 \\
\textbf{p-value} & \h0.9666 & \h0.0567 & \h0.4110 & \h0.9952 & \h0.9952 & \h0.9952 & \h0.9968 & \h0.5895 & 0.0000 & 0.0006 & \h0.9994 & \h0.9994 \\ \bottomrule
			\end{tabular}
		}
	\end{specialtable}

	We note in terms of forecasting accuracy (Tables~\ref{tab:sig_rmse} and~\ref{tab:sig_smape}) in each instance that the null hypothesis w.r.t. both predictands is rejected at the $\alpha = 0.01$ level of significance, except for LSTM, LGB, VARMAX, and MLR. This may seem contrary to previously presented results in this section, but the box plots in Figure~\ref{fig:box} explain this. Although the core and lower distributions may overlap (the box and bottom whisker) for the distributions in questions compared to MES-LSTM, the higher inaccuracies are only reported for the benchmarks, i.e. the upper whiskers for the benchmarks peak higher than MES-LSTM..
	
	For the prediction interval performance (Tables~\ref{tab:sig_mis} and~\ref{tab:sig_cov}) we note that we are able to reject the null hypothesis at the $\alpha = 0.01$ level of significance for MIS in all instances. Finally, we are unable to conclude that our model's Coverage Score is not outperformed except perhaps by DeepAR and VARMAX for \textit{tota; deaths}. Again, the t-test results are consistent with the results previously discussed.
	
	We also conduct a Diebold-Mariano (DM,~\cite{Die95}) test to compare the out-of-sample predictive skill of MES-LSTM to the benchmark models (Table~\ref{tab:dm_test}). The test is configured to use Mean Absolute prediction Error (MAPE,~\cite{Swam00}) and the null hypothesis is H$_0$: There is no significant difference in the accuracy of the competing forecasts. We reject the null hypothesis w.r.t. both predictands at the $\alpha = 0.01$ level of significance.
	
	\begin{specialtable}[htbp]
		\centering
		\caption{Diebold-Mariano test for Forecast accuracy (H$_0$: Benchmark Forecasts Outperform MES-LSTM.)}
		\label{tab:dm_test}
		\resizebox{1\columnwidth}{!}{%
			\begin{tabular}{@{}lrrrrrr|rrrrrr@{}}
				\toprule
				& \multicolumn{6}{c}{{\textbf{Total Cases}}} & \multicolumn{6}{c}{{\textbf{Total Deaths}}} \\ 
				& \textbf{LSTM} &
				\textbf{LGB} &
				\textbf{DeepAR} &
				\textbf{VARMAX} & \textbf{SARIMAX} & \textbf{MLR} & \textbf{LSTM} &
				\textbf{LGB} &
				\textbf{DeepAR} &
				\textbf{VARMAX} & \textbf{SARIMAX} & \textbf{MLR} \\ \bottomrule\toprule
				\textbf{statistic} & -19.817 & -277.776 & -144.279 & -12.665 & -16.160 & -8.048 & -5.567 & -374.711 & -112.500 & -7.066 & -11.817 & -43.647 \\
\textbf{p-value} & 0.000 & 0.000 & 0.000 & 0.000 & 0.000 & 0.000 & 0.000 & 0.000 & 0.000 & 0.000 & 0.000 & 0.000 \\ \bottomrule
			\end{tabular}
		}
	\end{specialtable}
	
	 We conclude this chapter with a deeper look into the effects of introducing more variability on our model's performance.
	
	\subsection{Effects of Variability on Model Performance}
	Our methodology suggests that the introduction of more variability into the data impacts both forecasting accuracy and prediction interval construction. We focus here specifically on MES-LSTM. We note from Figures~\ref{fig:map_forecast} and~\ref{fig:map_pi} that in most instances, the countries in which the predictions are least (most) accurate are also the countries in which the prediction intervals are widest (narrowest). This direct correlation reinforces the consistency of our model. We have normalized the MIS in Figure~\ref{fig:map_pi} before plotting the maps for ease of interpretation.
	
	\begin{figure}[htbp]
		\centering
		\begin{subfigure}{0.3\columnwidth}
			\includegraphics[trim=2.5cm 3cm 5.2cm 0.5cm, clip=true, width=\columnwidth]{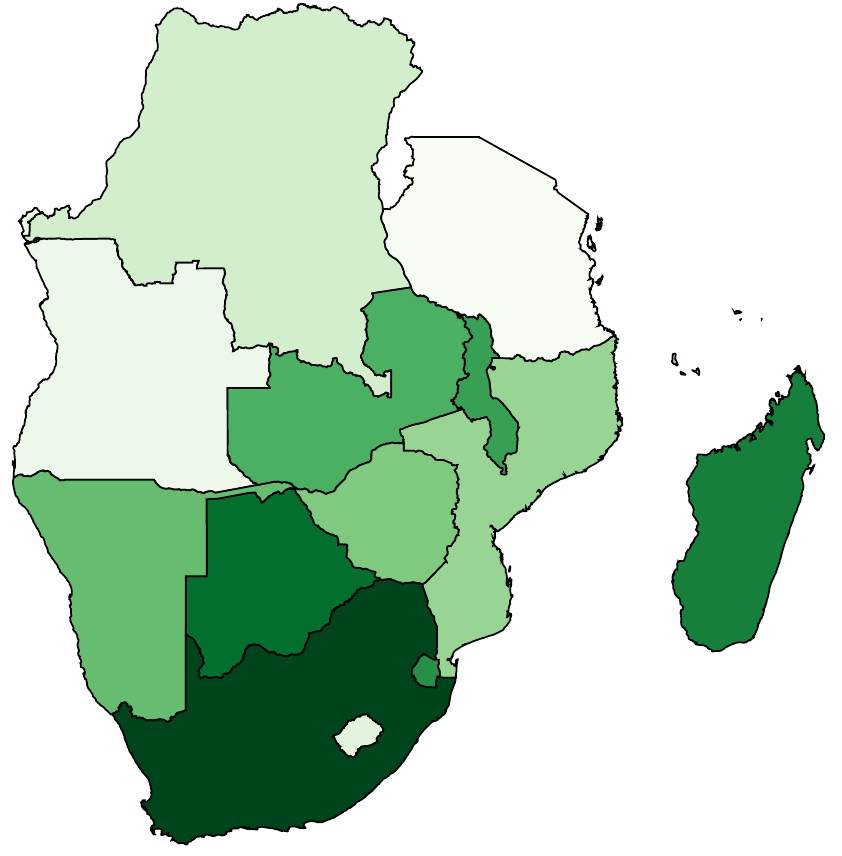}
			\caption{sMAPE for \textit{total cases}.}
			\label{fig:smape_cases_map}
		\end{subfigure}
		\begin{subfigure}{0.08\columnwidth}
			\includegraphics[trim=12.5cm 0cm 0cm 0cm, clip=true, width=\columnwidth]{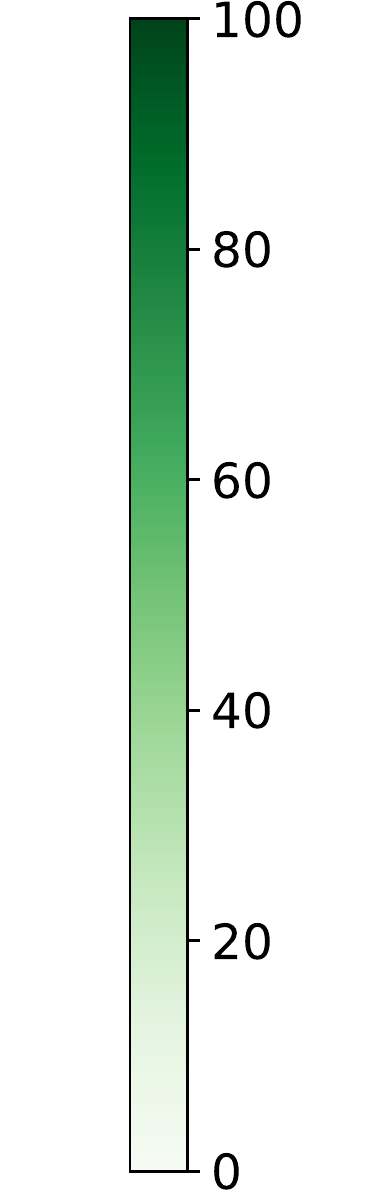}
		\end{subfigure}
		\begin{subfigure}{0.3\columnwidth}
			\includegraphics[trim=2.5cm 3cm 5.2cm 0.5cm, clip=true, width=\columnwidth]{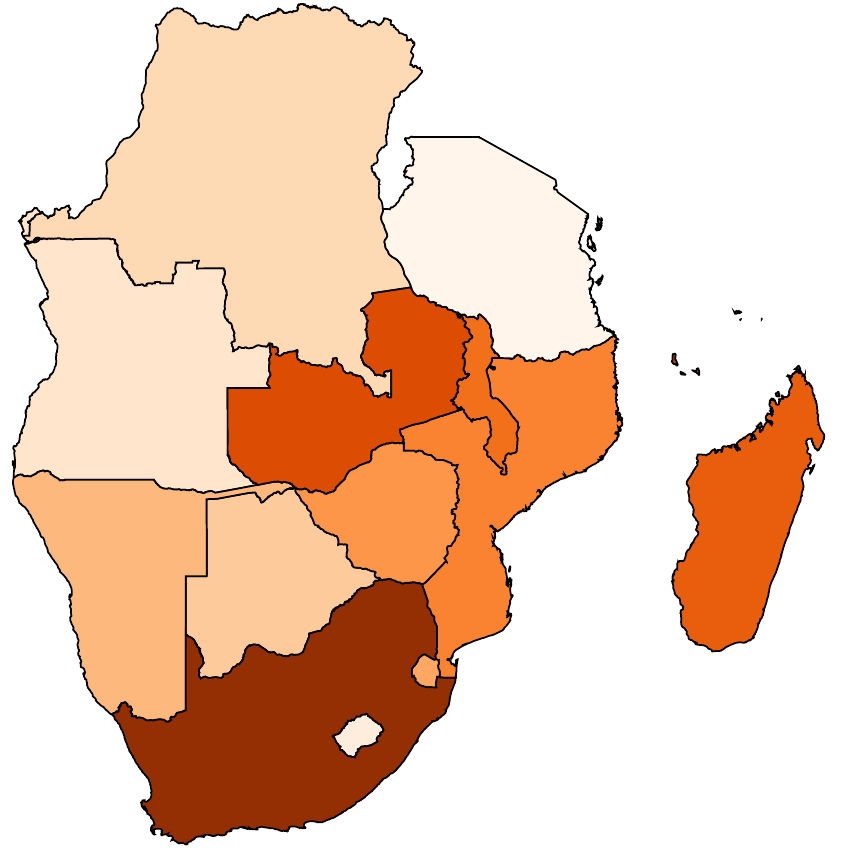}
			\caption{sMAPE for \textit{total deaths}.}
			\label{fig:smape_deaths_map}
		\end{subfigure}
		\begin{subfigure}{0.08\columnwidth}
			\includegraphics[trim=12.5cm 0cm 0cm 0cm, clip=true, width=\columnwidth]{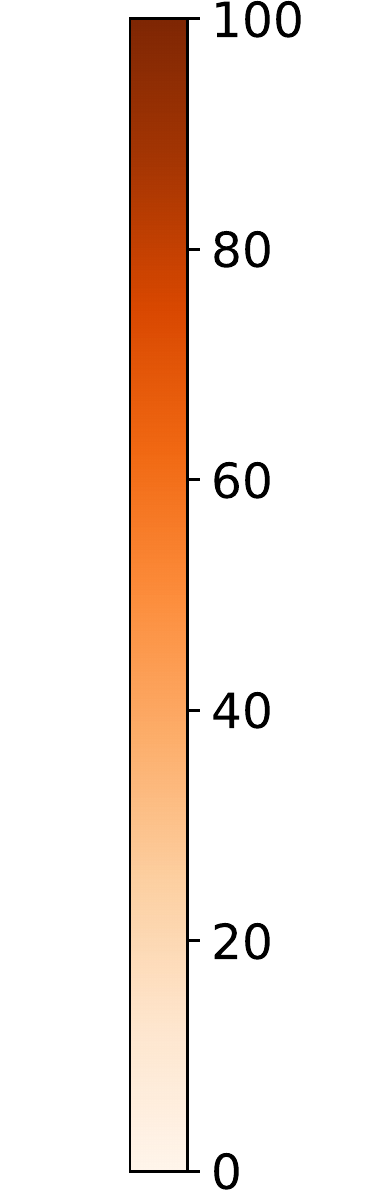}
		\end{subfigure}
		\caption{MES-LSTM Forecast Accuracy Ranked for Each Country in the SADC Region.\label{fig:map_forecast}}
	\end{figure}
	
	\begin{figure}[htbp]
		\begin{subfigure}{0.29\columnwidth}
			\includegraphics[trim=2.5cm 3cm 5.2cm 0.5cm, clip=true, width=\columnwidth]{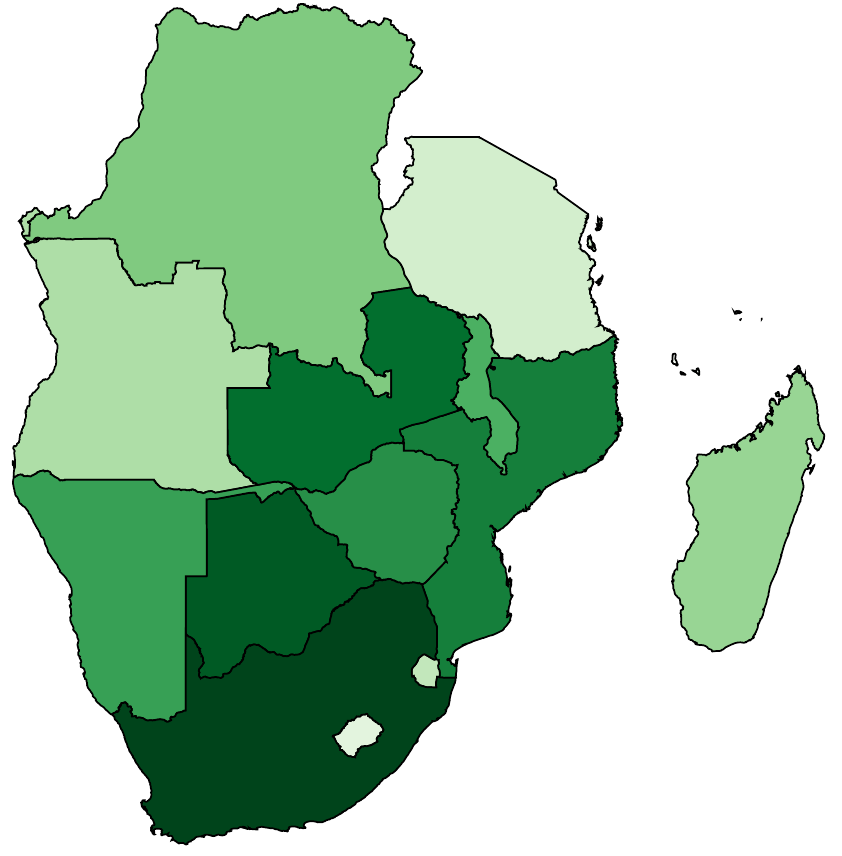}\caption{\textit{total cases} ($\alpha = 0.05$).\label{fig:mis_cases_map_05}} 
		\end{subfigure}
		\begin{subfigure}{0.29\columnwidth}
			\includegraphics[trim=2.5cm 3cm 5.2cm 0.5cm, clip=true, width=\columnwidth]{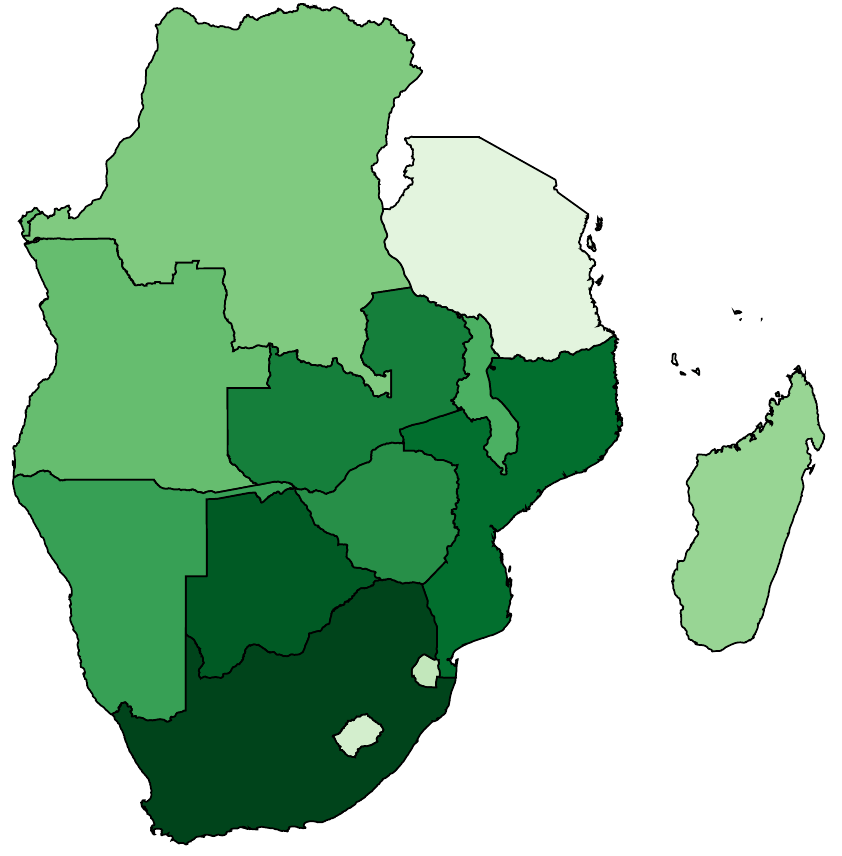}\caption{\textit{total cases} ($\alpha = 0.1$).\label{fig:mis_cases_map_1}}
		\end{subfigure}
		\begin{subfigure}{0.29\columnwidth}
			\includegraphics[trim=2.5cm 3cm 5.2cm 0.5cm, clip=true, width=\columnwidth]{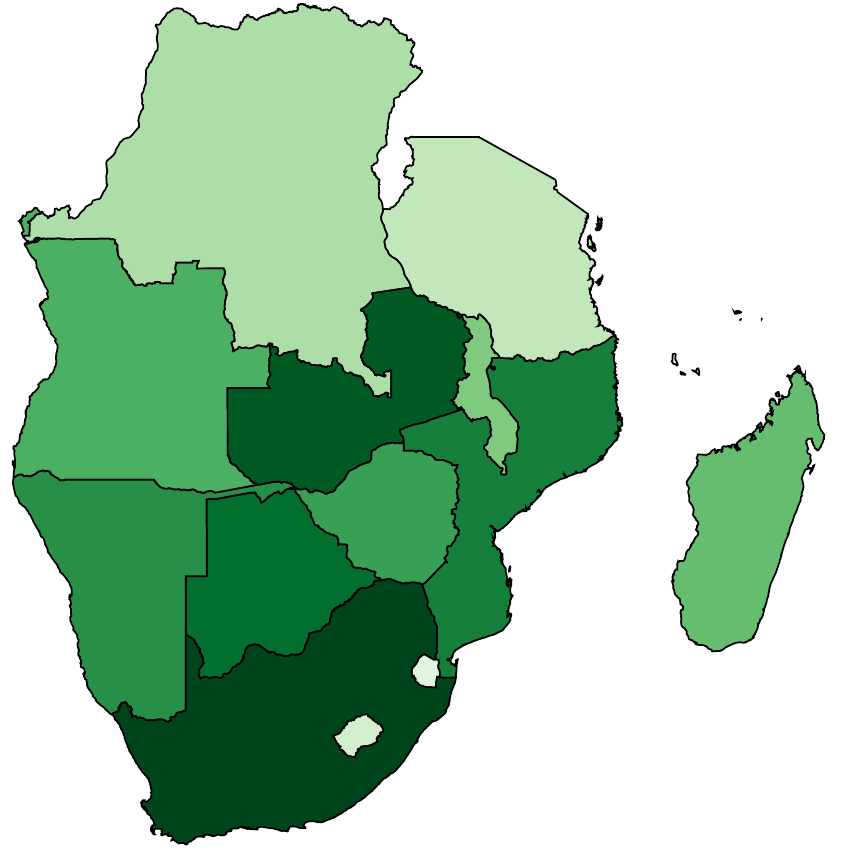}\caption{\textit{total cases} ($\alpha = 0.2$).\label{fig:mis_cases_map_2}}
		\end{subfigure}
		\begin{subfigure}{0.11\columnwidth}
			\includegraphics[trim=11.25cm 0cm 0cm 0cm, clip=true, width=\columnwidth]{Figures/map_cases_smape_1_legend.pdf}
		\end{subfigure}\\ \vfill
		\begin{subfigure}{0.29\columnwidth}
			\centering 
			\includegraphics[trim=2.5cm 3cm 5.2cm 0.5cm, clip=true, width=\columnwidth]{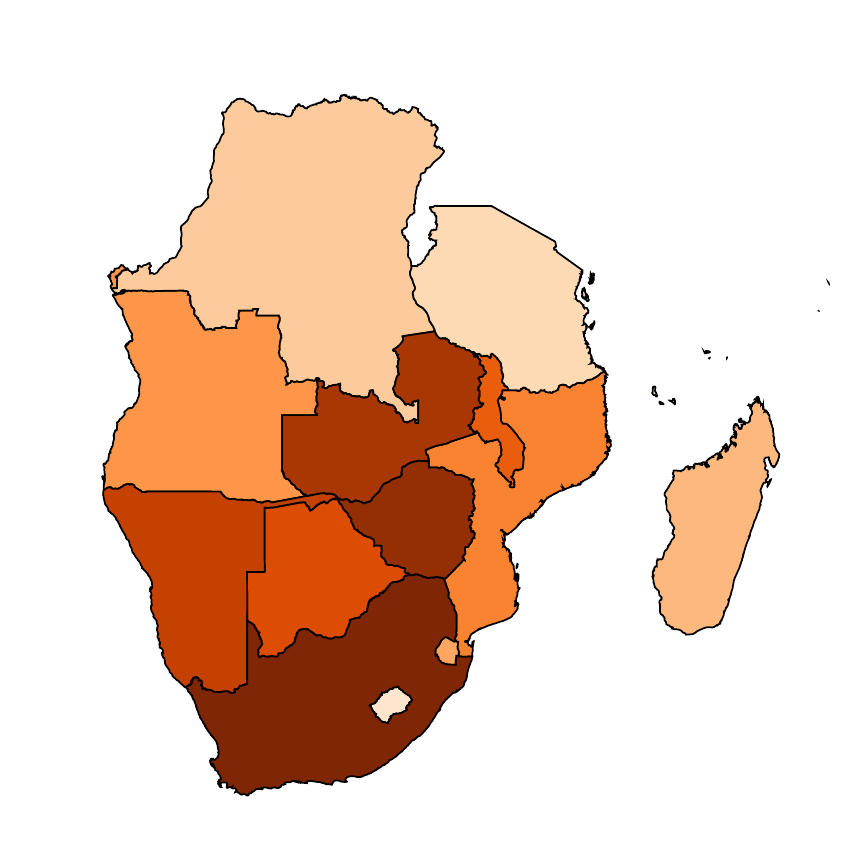}\caption{\textit{total deaths} ($\alpha = 0.05$).\label{fig:mis_deaths_map_05}}
		\end{subfigure}
		\begin{subfigure}{0.29\columnwidth}
			\centering 
			\includegraphics[trim=2.5cm 3cm 5.2cm 0.5cm, clip=true, width=\columnwidth]{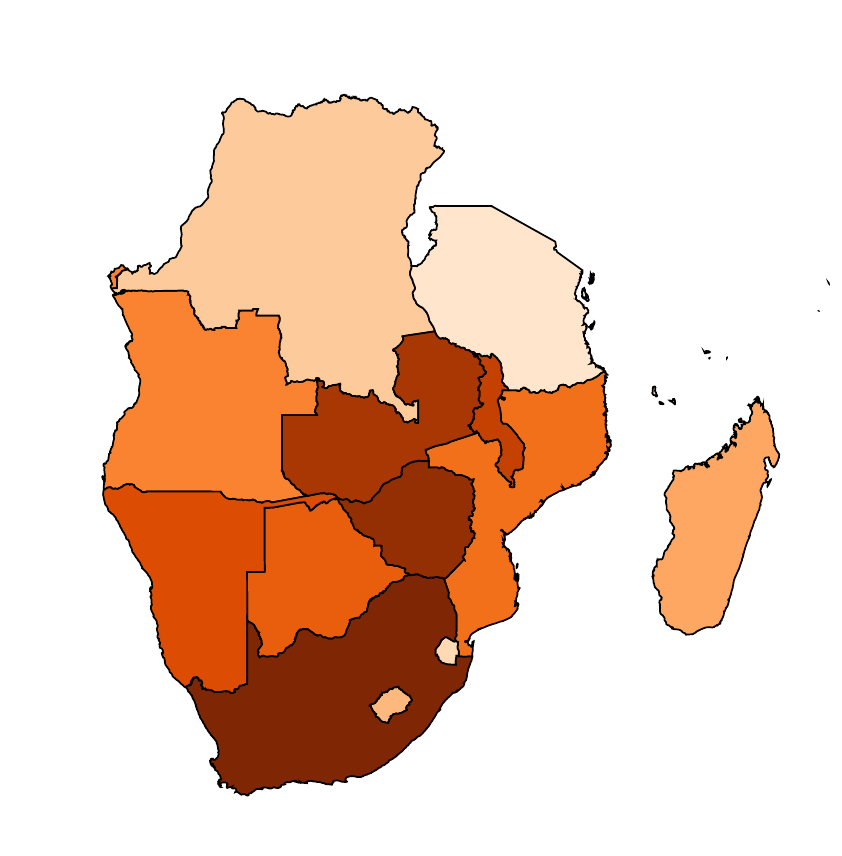}\caption{\textit{total deaths} ($\alpha = 0.1$).\label{fig:mis_deaths_map_1}}
		\end{subfigure}
		\begin{subfigure}{0.29\columnwidth}
			\includegraphics[trim=2.5cm 3cm 5.2cm 0.5cm, clip=true, width=\columnwidth]{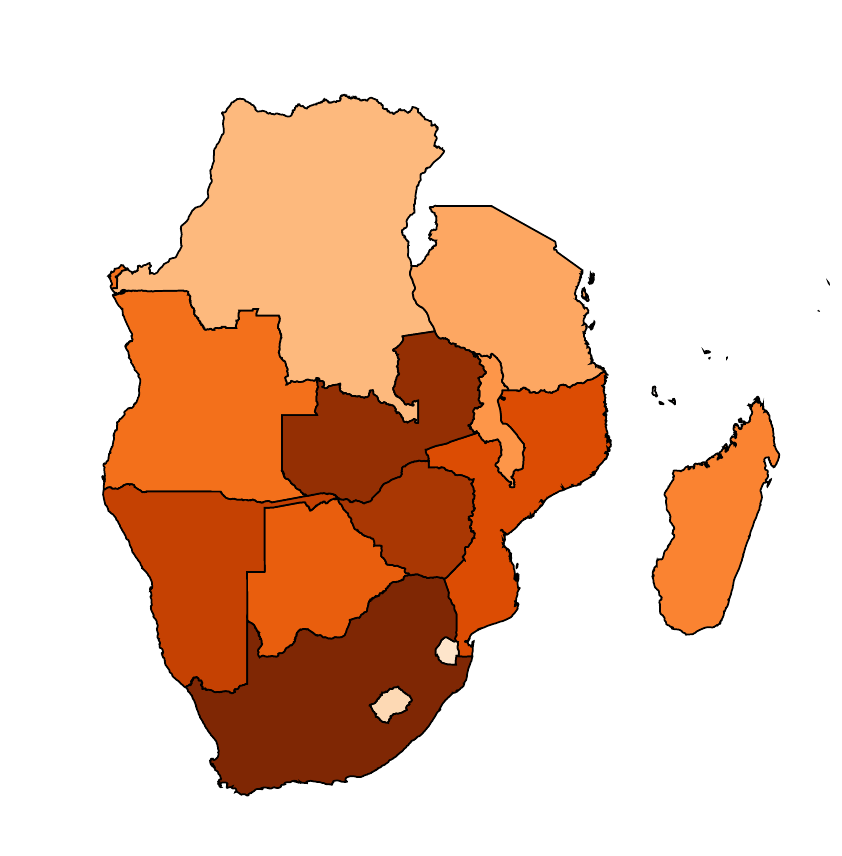}\caption{\textit{total deaths} ($\alpha = 0.2$).\label{fig:mis_deaths_map_2}}
		\end{subfigure}
		\begin{subfigure}{0.11\columnwidth}
			\includegraphics[trim=11.25cm 0cm 0cm 0cm, clip=true, width=\columnwidth]{Figures/map_deaths_smape_1_legend.pdf}
		\end{subfigure}
		\caption{MES-LSTM Prediction Interval Accuracy (normalized MIS) Ranked for Each Country in the SADC Region.\label{fig:map_pi}}
	\end{figure}
	
	In terms of Coverage, MES-LSTM is in instances outperformed by the benchmark methods. This Coverage is an area where the model can be improved. However, there are some crucial areas where the model improves on the skill of its counterparts. Our methodology suggests that the hybrid MES-LSTM is indeed able to outperform statistical methods and  deep learning techniques at both forecasting tasks and prediction interval construction for morbidity and mortality data with exogenous factors. In terms of the overall prediction error, there is a significant improvement over the benchmark models considered. Our model reports forecast consistently within a tight range for multiple independent trials. We also note that MES-LSTM offers the narrowest prediction intervals for all the predictands for all geographical regions at all the significance levels considered.
	
	\section{Conclusion}
	\label{sec:conclusion}
	We introduce a hybrid model, MES-LSTM, for multivariate prediction and forecast uncertainty quantification and apply it to morbidity and mortality datasets with exogenous factors. The univariate counterpart, Smyl's ES-RNN~\cite{Smyl20}, has been shown to perform well in the univariate case, outperforming both pure machine learning and pure statistical methods. We hypothesise that our multivariate extension also outperforms statistical and machine learning models at both forecasting tasks and constructing prediction intervals. With the methodology presented and the datasets considered, MES-LSTM significantly improves the skill of its classical probabilistic and pure deep learning counterparts.
	
	MES-LSTM shows consistent outperformance with forecast accuracy and the MIS of the prediction intervals constructed at all significance levels considered. There remains room for improvement when it comes to the coverage of the prediction intervals.
	
	In this paper we mostly limit our attention to the multivariate setting (except in Section~\ref{subsec:prelayer} where we use the univariate exposition as a building block before introducing our model). Future work may include running our model on univariate datasets as well. The benchmarks can also be applied with ease since both SARIMAX and VARMAX are univariate models if no exogenous inputs are declared, MLR is fit using OLS, and gradient boosting techniques can also be applied to univariate data.
	
	Applying the kind of techniques discussed in this paper to varied data with expedience is still a major limitation in related research. For example, the M5 Competitions although arguably among the most important forecast competitions globally, only considers retail units for one chain of supermarket. Future work may also include applying our model to more multivariate datasets from a broader cross-section of industries and applications beyond morbidity and mortality modeling.
	
	We motivate our choice of benchmark models in Section~\ref{subsec:bench}, but we also consider future work comparing MES-LSTM against more deep learning models such as convolutional, attention and transformer models. We also consider future applications where an adaptation or extension of our model can be applied, such as in multivariate anomaly detection.
	
	

	\vspace{6pt} 
	
	\supplementary{The following are available online at \url{https://github.com/zulucomputer/MES_LSTM}, code repository: MES-LSTM.}
	
	\authorcontributions{Conceptualization, T. Mathonsi and T. L. van Zyl; methodology, T. Mathonsi and T. L. van Zyl; software, T. Mathonsi; validation, T. Mathonsi; formal analysis, T. Mathonsi and T. L. van Zyl; investigation, T. Mathonsi and T. L. van Zyl; resources, T. Mathonsi; data curation, T. Mathonsi; writing---original draft preparation, T. Mathonsi; writing---review and editing, T. Mathonsi and T. L. van Zyl; visualization, T. Mathonsi and T. L. van Zyl; supervision, T. L. van Zyl; project administration, T. Mathonsi. All authors have read and agreed to the published version of the manuscript.}
	
	
	\conflictsofinterest{The authors declare no conflict of interest.} 
	
\end{paracol} 
\reftitle{References}


\externalbibliography{yes}
\bibliography{references_03.bib}

\end{document}